\RequirePackage[svgnames]{xcolor}

\documentclass[10pt,logo,letterpaper]{berkeley}

\usepackage[utf8]{inputenc} %
\usepackage[T1]{fontenc}    %
\usepackage{textcomp}
\usepackage[svgnames]{xcolor}
\usepackage{asymptote}
\usepackage[comma,authoryear,compress]{natbib}
\usepackage{wrapfig}
\usepackage{hyperref}
\usepackage{algorithm}
\usepackage{titletoc}
\usepackage{algorithmicx}
\usepackage{wrapfig}
\usepackage{algpseudocode}
\usepackage{microtype}
\usepackage{graphicx}
\expandafter\def\csname ver@subfig.sty\endcsname{}
\usepackage{booktabs} %
\usepackage{float}
\usepackage{bigstrut}
\usepackage{tikz}
\usetikzlibrary{tikzmark}
\makeatletter
\newcommand*\myfontsize{%
  \@setfontsize\myfontsize{7}{8}%
}
\makeatother
\definecolor{myred}{rgb}{0.7, 0.3, 0.0}
\definecolor{myblue}{HTML}{054488}
\definecolor{mygreen}{HTML}{056b34}

\newcolumntype{R}[1]{>{\raggedleft\let\newline\\\arraybackslash\hspace{0pt}}m{#1}}

\usetikzlibrary{intersections}

\definecolor{darkgreen}{rgb}{0.0, 0.42, 0.24}

\usepackage{booktabs}
\usepackage{multirow}
\usepackage{colortbl}
\usepackage{xcolor}
\usepackage{booktabs}
\usepackage{multirow}
\definecolor{bestcolor}{RGB}{144, 238, 144}  
\definecolor{secondcolor}{RGB}{173, 216, 230}  
\definecolor{headercolor}{RGB}{240, 240, 240}  
\usepackage{amsthm}
\usepackage{nicefrac}
\usepackage{subcaption}
\usepackage{caption}
\usepackage{graphicx}
\usepackage{cleveref}
\IfFileExists{bxcoloremoji.sty}{\usepackage{bxcoloremoji}}{\providecommand{\coloremojicode}[1]{}}
\usepackage{listings}
\usepackage{float}
\usepackage{longtable}
\usepackage{booktabs}
\usepackage{multirow}
\usepackage{xcolor}
\usepackage{colortbl} 

\definecolor{highlight}{HTML}{F2F2F2}
\lstdefinestyle{Python}{
    language=Python,
    basicstyle=\ttfamily\footnotesize,
    keywordstyle=\color{blue}\bfseries,
    commentstyle=\color{green},
    stringstyle=\color{red},
    numberstyle=\tiny\color{gray},
    showstringspaces=false,
    frame=single,
    breaklines=true,
    backgroundcolor=\color{lightgray!20}
}

\usepackage{hyperref}       %
\usepackage{url}            %
\usepackage{booktabs}       %
\usepackage{nicefrac}       %
\usepackage{microtype}      %
\usepackage{graphicx}
\usepackage{subcaption} 
\usepackage{wrapfig}
\usepackage{lipsum}
\usepackage{enumitem}
\usepackage{stackengine}
\usepackage[font=small,labelfont=bf]{caption}
\usepackage{color}
\usepackage{adjustbox}
\usepackage{tikz}
\usepackage{tcolorbox}
\tcbuselibrary{breakable, skins}
\usepackage{rotating}
\usepackage{makecell}
\usepackage{colortbl}    
\usepackage{multirow}    
\usepackage{pifont}      
\usepackage{xcolor}      
\usepackage{array}       
\usepackage{amsmath}
\usepackage{amsfonts}       %
\usepackage{xspace}
\usepackage{mathtools}
\usepackage{xcolor}
\usepackage{colortbl}
\usepackage{tabularx}
\usepackage{tcolorbox}
\newcommand{\cmark}{\ding{51}}   
\newcommand{\xmark}{\ding{55}}   
\definecolor{oursblue}{RGB}{230,240,255} 

\definecolor{blanchedalmond}{rgb}{1.0, 0.92, 0.8}
\definecolor{carmine}{rgb}{0.59, 0.0, 0.09}
\definecolor{lightblue}{rgb}{0.22,0.45,0.70}%

\renewcommand{\mathbf}{\boldsymbol}

\makeatletter
\def\Ddots{\mathinner{\mkern1mu\raise\p@
\vbox{\kern7\p@\hbox{.}}\mkern2mu
\raise4\p@\hbox{.}\mkern2mu\raise7\p@\hbox{.}\mkern1mu}}
\makeatother

\definecolor{amaranth}{rgb}{0.9, 0.17, 0.31}
\definecolor{antiquebrass}{rgb}{0.8, 0.58, 0.46}
\definecolor{antiquefuchsia}{rgb}{0.57, 0.36, 0.51}
\definecolor{chromeyellow}{rgb}{0.31, 0.47, 0.26}

\newtcolorbox{AIbox}[2][]{aibox,title=#2,#1}
\definecolor{lightblue}{rgb}{0.22,0.45,0.70}%
\definecolor{Gray}{gray}{0.95}
\definecolor{Cornsilk}{rgb}{1.0, 0.97, 0.86}
\usepackage{enumitem}
\usepackage{booktabs}
\usepackage{multirow}
\usepackage{colortbl}
\usepackage{xcolor}
\definecolor{highlightcolor}{RGB}{235, 250, 235}
\definecolor{avgcolor}{gray}{0.92}

\newcolumntype{g}{>{\columncolor{avgcolor}}c}

\makeatother

\newcommand{\imp}[1]{\textcolor{darkgreen}{\scriptsize{\textbf{(#1)}}}}
\definecolor{myred}{rgb}{0.7, 0.3, 0.0}
\definecolor{myblue}{HTML}{054488}
\definecolor{mygreen}{HTML}{056b34}
\definecolor{myorange}{HTML}{ff8800}
\definecolor{mypurple}{HTML}{8400ff}
\definecolor{mypink}{HTML}{f7acb9}

\definecolor{myred}{rgb}{0.7, 0.3, 0.0}
\definecolor{myblue}{HTML}{054488}
\definecolor{mygreen}{HTML}{056b34}
\definecolor{tiktokpink}{HTML}{E91E63}
\definecolor{tiktokpurple}{HTML}{673AB7}
\definecolor{tiktokgray}{HTML}{9E9E9E}

\newcommand{\mytitle}{CaveAgent: Transforming LLMs into Stateful Runtime Operators}

\usepackage{amssymb}
\usepackage{algorithm}
\usepackage{algpseudocode}
\usepackage{booktabs}
\usepackage{graphicx}
\usepackage{listings}
\usepackage{xcolor}
\usepackage{pgfplots}
\pgfplotsset{compat=1.17}

\title{\mytitle}

\runningtitle{\mytitle}

\author{\small
  Maohao Ran$^{* 1,2}$,
  Zhenglin Wan$^{* 3}$,
  Cooper Lin$^{4}$,
  Yanting Zhang$^{1}$,
  Hongyu Xin$^{1}$,
  Hongwei Fan$^{7}$,
  Yibo Xu$^{1}$,
  Beier Luo$^{4}$,
  Yaxin Zhou$^{5}$,
  Wangbo Zhao$^{3}$,
  Lijie Yang$^{8}$,
  Lang Feng$^{6}$,
  Fuchao Yang$^{6}$,
  Jingxuan Wu$^{9}$,
  Yiqiao Huang$^{10}$,
  Chendong Ma$^{2}$,
  Yusen Huang$^{11}$,
  Dailing Jiang$^{2}$,
  Jianbo Deng$^{1}$,
  Sirui Han$^{1}$,
  Yang You$^{3}$,
  Bo An$^{6}$,
  Yike Guo$^{1}$,
  Jun Song$^{1,2\dagger}$
}
\affil{\normalsize{$^1$HKUST \quad $^2$HKBU \quad $^3$NUS \quad $^4$HKU \quad $^5$CMU \quad 
  $^6$NTU, Singapore \\  $^7$Imperial College London \quad $^8$Princeton \quad $^9$UNC, Chapel Hill \quad $^{10}$Harvard \quad $^{11}$CUHK-Shenzhen}
}

\correspondingauthor{$^*$ Joint first author \& Equal Contribution.\\$\dagger$ Corresponding to junsong@hkbu.edu.hk. Work conducted at the Hong Kong Generative AI Research and Development Center (HKGAI), led by HKUST.}

\graphicspath{{./}{figures/}{figures/CaveAgent/}}
\providecommand{\cmark}{\ding{51}}
\providecommand{\xmark}{\ding{55}}
\definecolor{darkgreen}{rgb}{0.0, 0.42, 0.24}
\definecolor{promptbg}{RGB}{245,245,245}
\definecolor{promptframe}{RGB}{200,200,200}
\definecolor{bestcolor}{RGB}{144, 238, 144}
\definecolor{secondcolor}{RGB}{173, 216, 230}
\definecolor{headercolor}{RGB}{240, 240, 240}
\definecolor{highlight}{HTML}{F2F2F2}
\definecolor{oursblue}{RGB}{230,240,255}
\lstdefinestyle{Python}{language=Python, basicstyle=\ttfamily\footnotesize,
  keywordstyle=\color{blue}\bfseries, commentstyle=\color{green}, stringstyle=\color{red},
  numberstyle=\tiny\color{gray}, showstringspaces=false, frame=single, breaklines=true,
  backgroundcolor=\color{lightgray!20}}

\begin{document}

\begin{abstract}
\vspace{-0.6cm}
\paragraph{Abstract:} LLM-based agents are increasingly capable of complex task execution, yet current agentic systems remain constrained by text-centric paradigms that struggle with long-horizon tasks due to fragile multi-turn dependencies and context drift. We present CaveAgent, a framework that shifts LLM tool use from ``LLM-as-Text-Generator'' to ``LLM-as-Runtime-Operator.'' CaveAgent introduces a dual-stream architecture: a \textit{semantic stream} for lightweight reasoning and a \textit{runtime stream} backed by a persistent Python environment for stateful execution. Rather than treating the LLM's text context as the primary workspace, CaveAgent elevates the persistent runtime as the central locus. Beyond leveraging code generation to resolve interdependent sub-tasks (e.g., loops, conditionals) in a single step, CaveAgent introduces Stateful Runtime Management: it injects, manipulates, and retrieves complex Python objects (e.g., DataFrames, database connections) that persist across turns, unlike existing code-based approaches that remain text-bound. CaveAgent further provides a runtime-integrated skill management system that extends the Agent Skills open standard, enabling ecosystem interoperability through executable skill injections. This persistence mechanism serves as a high-fidelity external memory that reduces context drift in multi-turn interactions and preserves processed data for downstream applications with less information loss. Evaluations on Tau$^2$-bench and the Berkeley Function Calling Leaderboard (BFCL) across six state-of-the-art LLMs demonstrate consistent improvements in 11 out of 12 settings, with gains up to +13.5\% success rate on multi-turn retail tasks. On BFCL, the three open-source models we evaluate all reach 94.0--94.7\% under CaveAgent, comparable to closed-source Claude Sonnet 4.5 (94.4\%) and Gemini 3 Pro (94.3\%) and exceeding GPT-5.1 (89.6\%) under their native function-calling protocols; the 30B Qwen3-Coder reaching 94.4\% suggests the function-calling protocol is a key performance bottleneck alongside model scale. Token efficiency studies show 28.4\% reduction in total token consumption and up to 51\% token reduction on data-intensive tasks relative to the best baseline. The accessible runtime state further provides programmatically verifiable feedback, enabling automated evaluation and reward signal generation without human annotation and establishing a structural foundation for future research in Reinforcement Learning with Verifiable Rewards (RLVR).

\vspace{0.2cm}

\coloremojicode{1F4C5} \textbf{Date}: Jun 25, 2026 (v3)

\coloremojicode{1F4BB} \textbf{Code}: \href{https://github.com/acodercat/cave-agent}{https://github.com/acodercat/cave-agent} \qquad


\coloremojicode{1F4E7} \textbf{Main Contact}: Zhenglin Wan~(\href{mailto:vanzl@u.nus.edu}{vanzl@u.nus.edu}), Jun Song~(\href{mailto:junsong@hkbu.edu.hk}{junsong@hkbu.edu.hk})

\end{abstract}

\maketitle

\begin{figure}[htbp]
    \centering
    \adjustbox{cfbox=gray!50 0.5pt}{\includegraphics[width=\textwidth]{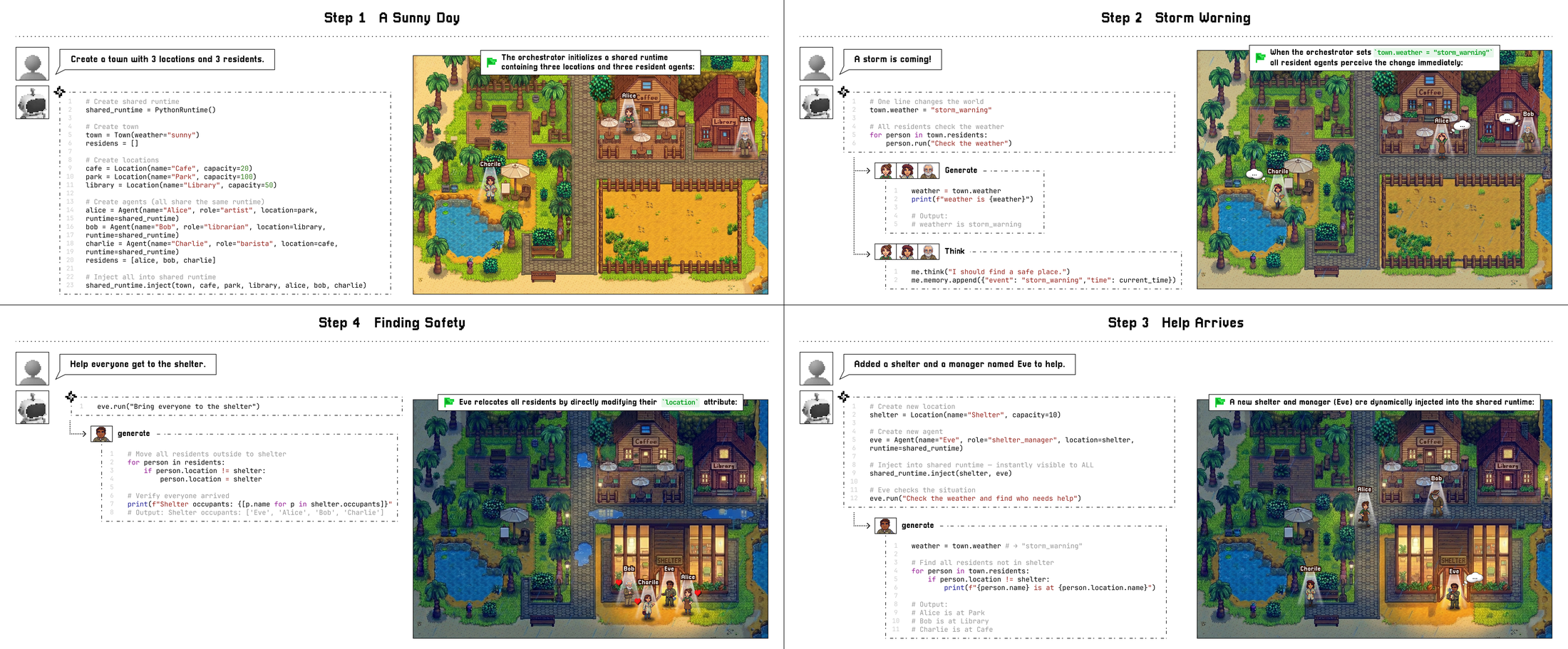}}
    \caption{Town Simulation: a toy example for Stateful Runtime-Mediated Multi-Agent Collaboration. Multiple resident agents share a single Python runtime; when the meta-agent modifies a global object (e.g., the weather entity), all residents observe the change through direct reference rather than message passing.}
    \label{Town}
\end{figure}

\section{Introduction}

Large Language Models (LLMs) have demonstrated strong knowledge acquisition and reasoning capabilities across diverse natural language processing tasks. Building on these capabilities, tool-integrated reasoning (TIR) enables LLM agents to interact with external tools and APIs in a multi-turn manner\footnote{In this paper, we use tool use and function calling interchangeably.} \citep{lu2023chameleon, shen2023hugginggpt, patil2024gorilla, qu2025tool}, expanding their information access and solution space. This has extended LLM agents to various domains, including scientific discovery \citep{boiko2023emergent, bran2024chemcrow}, mathematical problem-solving \citep{gao2023pal, chen2022program}, Web GUI navigation \citep{zhou2023webarena, yao2022webshop}, and robotics \citep{driess2023palme, brohan2023rt2}.

Despite this progress, the conventional protocol for tool use requires LLMs to conform to predefined JSON schemas and generate structured JSON objects containing precise tool names and arguments \citep{qin2023toolllm, openai2023gpt4}. For example, to retrieve stock data, the model must synthesize a JSON string like \texttt{\{"tool": "get\_stock", "params": \{"ticker": "AAPL", "date": "today"\}\}}, requiring exact adherence to syntax and field constraints. This imposes three architectural limitations: \textbf{1) Rigid Control Flow:} each turn executes a single tool call (or parallel batch) and serializes output back to context, introducing latency for tasks requiring sequential orchestration \citep{wu2023autogen, shen2023hugginggpt}; \textbf{2) Statelessness:} each call is an isolated transaction with no persistent state across turns, so intermediate results must be serialized back into text, causing token overhead for complex data structures and cascading error propagation \citep{qiao2023taskweaver, kim2023language}; and \textbf{3) Limited Composability:} JSON schemas express flat function signatures and cannot natively represent loops, conditionals, or variable dependencies, forcing multi-step logic into error-prone multi-turn dialogue \citep{kim2023language, wang2024executable}.

While recent works attempt to address these issues with code-based tool use \citep{wang2024executable, yang2024if}, they predominantly adopt a \textit{process-oriented} paradigm where the runtime state remains \textbf{internalized and text-bound}. This creates a ``textualization bottleneck'': variables are accessible to external systems only through text output, requiring serialization into text strings (e.g., printing a DataFrame) to communicate with the user \citep{wang2024executable, yao2022react}. This prevents the direct input and output of structured, manipulatable objects, making it inefficient or impossible to handle complex non-textual data (e.g., large datasets, videos) \citep{qiao2023taskweaver} and interact with downstream tasks. To address these limitations:

\begin{tcolorbox}[colback=blue!5!white, colframe=blue!75!black, boxrule=0.6pt,
left=3pt, right=3pt, top=2pt, bottom=2pt]
    \emph{We aim to build a system that utilizes Python's "everything is an object'' philosophy to enable full \textit{Object-Oriented} function calling and interaction, delegating context engineering to a \textbf{persistent} runtime and allowing the direct injection and retrieval of high-fidelity objects without serialization loss, thereby fully leveraging the strong code-generation capability of LLMs, and enabling modular distribution of executable tool capabilities through a portable Agent-Skill standard.}
\end{tcolorbox}

We present \textbf{CaveAgent}\footnote{Code and data are publicly available at \url{https://github.com/acodercat/cave-agent}.}, an open-source framework that introduces the concept of \textbf{Stateful Runtime Management} for LLM agents, shifting code-based tool use from ``process-oriented function calling'' to persistent ``object-oriented state manipulation.'' CaveAgent operates on a \textbf{dual-stream architecture} with two distinct streams: a \textbf{semantic stream} for reasoning and a \textbf{runtime stream} for state management and code execution. This represents an architectural alternative to the conventional design: whereas existing agents treat the LLM's semantic context as the primary workspace with external tools as auxiliary, CaveAgent elevates the persistent runtime as the primary locus of computation and state, with the semantic stream serving as a lightweight orchestrator that generates code to manipulate it.

By injecting complex data structures (e.g., graphs, DataFrames) directly into the runtime as persistent objects, CaveAgent achieves a form of \textbf{context engineering}: the agent manipulates high-fidelity data via concise variable references, decoupling storage from the limited context window. Any intermediate result (e.g., DataFrames, planning trees, or key metadata) can be stored in persistent variables that the agent actively retrieves for later use or downstream applications (code as action, state as memory). This reduces progressive context degradation in multi-turn interactions \citep{catastrophic1, catastrophic2}, enables context compression, and provides error-free recall through the runtime serving as an external memory.

The persistent environment also enables \textbf{few-step resolution of complex logical dependencies} by using code to interact with multiple interdependent tools, allowing the agent to compose workflows (e.g., data filtering followed by analysis) in a few turns rather than through error-prone multi-round function calling \citep{wang2024mint, qin2023toolllm}. The runtime's transparency makes agent behavior fully verifiable, supporting checks on both intermediate programmatic states and final output objects of any data type, thereby enabling fine-grained reward signals for Reinforcement Learning.

Finally, CaveAgent supports \textbf{artifact handoff without information loss} by returning native Python objects rather than text representations, enabling direct use in downstream tasks such as UI rendering, visualization, and structured validation. The runtime can be serialized and reloaded, preserving the agent's complete state across sessions. This transforms the LLM from an isolated text generator into a stateful, interoperable computational component within broader software ecosystems.

The function-calling paradigm in CaveAgent also extends beyond single-agent capabilities to enable \textbf{Runtime-Mediated Multi-Agent Coordination}, one of several capabilities summarized in Figure~\ref{application}. Unlike conventional frameworks where agents coordinate via lossy text message passing \citep{li2023camel,generative_agents}, CaveAgent enables agents to interact through direct state manipulation. A supervisor agent can programmatically inject variables into a sub-agent's runtime to alter its environment or task context without ambiguous natural language instructions. Multiple agents can also operate on a unified shared runtime, achieving implicit synchronization: when one agent modifies a shared object (e.g., updating a global ``weather'' entity in a town simulation), the change is immediately visible to all peers through direct reference. This enables multi-agent collaboration through state-mediated coordination as an alternative to message passing (qualitative case studies in Appendix~\ref{app_multiagent}, including a town-simulation walkthrough; rigorous quantitative evaluation is left for future work). We summarize our contributions as follows:

\begin{figure}[t]
    \centering
    \includegraphics[width=\linewidth]{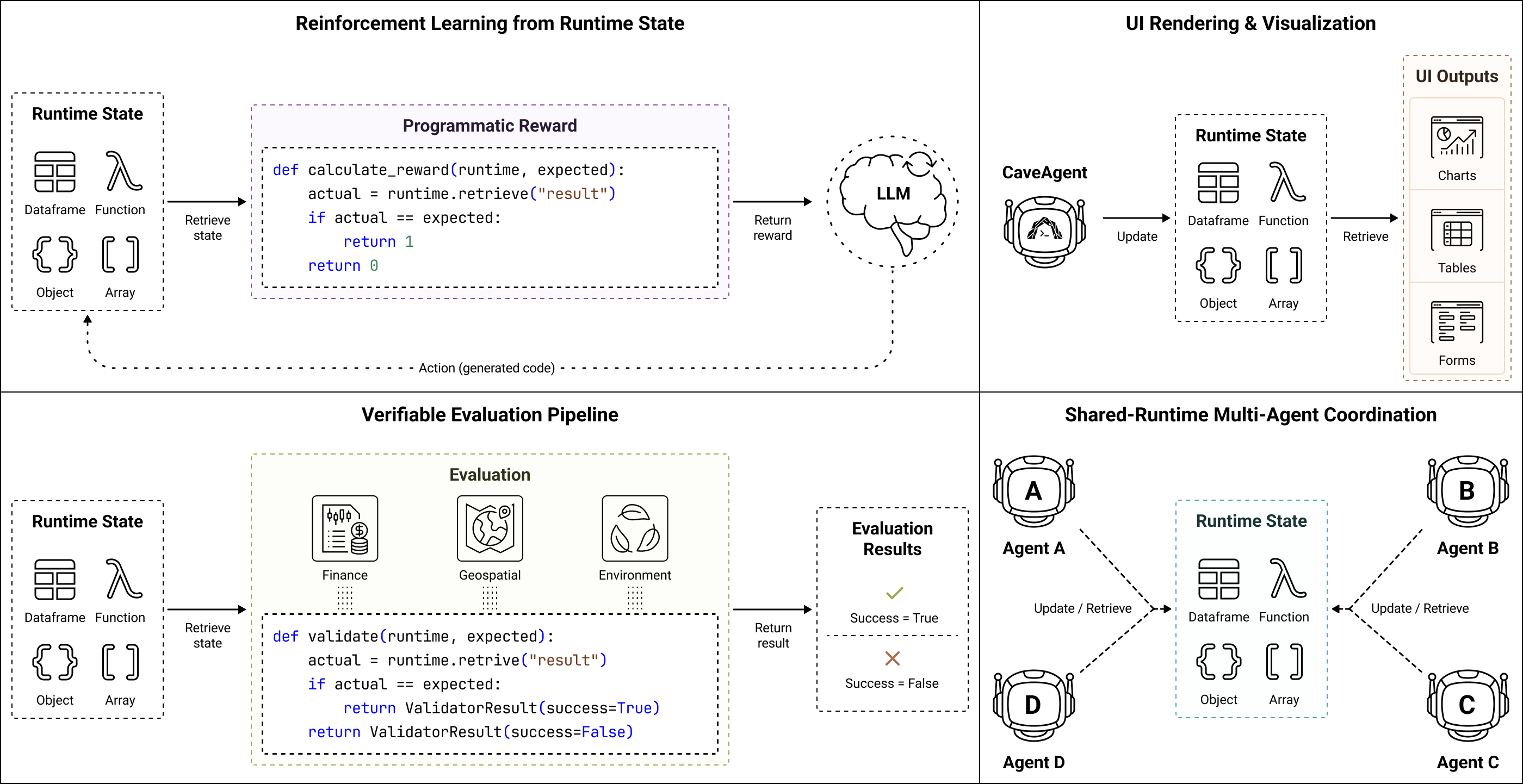}
    \caption{Key Advantages of CaveAgent}
    \label{application}
\end{figure}

\begin{itemize}[itemsep=0.35cm]
    \item We introduce \textbf{CaveAgent}, a tool-use framework built on \textbf{Stateful Runtime Management}. CaveAgent shifts the paradigm from process-oriented function calling to persistent, object-oriented state management. It achieves context compression and context-grounded memory recall by delegating context engineering to a persistent runtime, eliminating the token overhead and precision loss of textual serialization while enabling few-step resolution of logically interdependent tasks. CaveAgent also introduces \textbf{runtime-integrated skill management} that extends the Agent Skills open standard: because the architecture treats tools as first-class Python objects, skills can deliver executable artifacts (functions, variables, and type definitions) directly into the runtime upon activation, unifying tool registration and skill distribution into a single mechanism.

    \item The framework's programmatic inspectability is a \textit{structural} property of the architecture: runtime state is deterministically accessible at any point, providing automated evaluation and fine-grained reward signals for \textbf{Reinforcement Learning with Verifiable Rewards (RLVR)} without requiring subjective human annotation.

    \item We evaluate CaveAgent on standard benchmarks (e.g., Tau$^2$-Bench) and provide case studies across various domains including geospatial analysis (Appendix~\ref{app:geospatial}) and AutoML multi-agent coordination (Appendix~\ref{app:automl}). We also provide qualitative case studies suggesting how the same runtime substrate can support \textbf{Stateful Runtime-Mediated Multi-Agent Coordination}, leaving rigorous quantitative evaluation as a direction for future work.
\end{itemize}


\section{Background and Related Work}
The recent surge of autonomous LLM-driven agents has been comprehensively surveyed in the context of distributed AI for industrial deployment \citep{piccialli2025agentai}, and concrete domain deployments --- such as CDAFlow's stateful clinical decision-making framework \citep{hou2026cdaflow} --- underscore a practical demand for agents that can reliably manage state across multi-step workflows. These trends motivate renewed attention to the architectural choices behind state management, memory, and tool orchestration. We review four threads bearing directly on CaveAgent: tool learning, code-based action, context management, and multi-agent coordination.

\subsection{Tool Learning \& Function Calling}
The foundational approach to LLM tool use relies on a JSON-centric paradigm. The ReAct framework \citep{yao2022react} established the influential thought-action-observation loop, enabling LLMs to interleave reasoning with tool invocation. Building on this, JSON-Schema function calling \citep{patil2024gorilla, qin2023toolllm} constrains action generation to structured schemas, formalized by GPT-4 Function Calling and widely adopted by frameworks such as AutoGen \citep{wu2023autogen}. This schema-based approach has proven remarkably effective in practice, powering the majority of today's deployed agent systems with reliable type checking and standardized interfaces. Constrained decoding methods like xGrammar \citep{dong2024xgrammar} further enforce syntactically valid JSON at low overhead. Despite this success, JSON-based function calling faces inherent architectural trade-offs when applied to complex, multi-step tasks: as a static interchange format, JSON lacks native control flow; its verbose syntax incurs token overhead \citep{wang2024executable}; and each call operates as an isolated transaction, requiring all intermediate state to be serialized back into text \citep{packer2023memgpt,wang2024mint}.

\subsection{Code as Action \& Programmatic Reasoning}
The ``Code as Action'' paradigm represents a significant advance by using executable Python as a unified medium for reasoning and tool invocation. \citet{wang2024executable} proposed \textit{CodeAct}, which demonstrated that replacing JSON payloads with Python code reduces multi-turn overhead by up to 30\% and improves task success rates by 20\%. This paradigm leverages the Turing-complete nature of code to express loops, conditionals, and variable dependencies. The paradigm extends to domain-specific reasoning: \textit{ViperGPT} \citep{suris2023vipergpt} composes vision modules into executable subroutines, while \textit{Program of Thoughts} \citep{chen2022program} and \textit{PAL} \citep{gao2023pal} delegate arithmetic and symbolic logic to a Python interpreter, and \citet{bai2025collaboration} use intelligent agents to enrich the prompts driving LLM code generation. While CodeAct's persistent runtime is a key enabler, its interface remains \textit{text-bound}: intermediate states are communicated to external systems only via \texttt{print} output, and external data must be loaded through file I/O \citep{wang2024executable}. This makes it difficult to inject pre-existing Python objects (e.g., in-memory DataFrames, trained models) or retrieve runtime objects for downstream use without serialization loss \citep{liu2024lost,packer2023memgpt}.

\subsection{Context Management \& Stateful Architectures}
\citet{packer2023memgpt} introduced \textit{MemGPT}, an OS-inspired virtual context management system with tiered memory for long-horizon tasks. \citet{qiao2023taskweaver} proposed \textit{TaskWeaver}, a code-first framework preserving data structures across turns. However, existing approaches rely on RAG or textual summarization, inherently \textbf{lossy} methods that strip complex runtime objects of structural integrity and executable properties. CaveAgent uses \textit{Variable Injection} to treat the Python runtime itself as high-fidelity external memory, allowing variables to persist in their native object form without re-tokenization overhead.

\subsection{Multi-Agent Coordination}
\citet{li2023camel} proposed \textit{CAMEL} for role-playing cooperation; \citet{qian2023chatdev} introduced \textit{ChatDev} with chat-chain workflows; \citet{hong2023metagpt} developed \textit{MetaGPT} encoding SOPs into prompts; and \citet{saadaoui2025coordinated} more recently studied coordinated LLM multi-agent systems for collaborative question--answer generation. Although these frameworks differ in role assignment and orchestration patterns, they share a common substrate: agents communicate via \textbf{text-based message passing}, which introduces serialization bottlenecks when complex state must be transferred between agents. A separate, longer-running line of research on intelligent-agent architectures predates the LLM era --- for example, fuzzy-BDI agent models for cyber-physical systems \citep{karaduman2024impact} --- and offers complementary perspectives on stateful, autonomous reasoning, although it does not directly address the text-serialization issue we focus on here. CaveAgent instead enables \textit{Runtime-Mediated State Flow}: agents collaborate by directly injecting and retrieving variables in a shared runtime, shifting coordination from ``communication by talking'' to ``communication by shared state.''

Figure~\ref{evolution} illustrates the evolution of architectural approaches to agentic tool use that motivates our work.

\subsection{Comparison with Related Stateful Agent Frameworks}
\label{sec:framework_comparison}

Table~\ref{tab:framework_comparison} positions CaveAgent against three closely-related frameworks. CodeAct \citep{wang2024executable} shares CaveAgent's persistent code-execution kernel, but interaction with the runtime is one-way and text-bound: external data must be loaded via file I/O within generated code, and intermediate state is surfaced through stdout or text-formatted return values rather than as native Python objects. CaveAgent's three distinctive advances over CodeAct are: \textit{(i)} bidirectional \texttt{inject()}/\texttt{retrieve()} APIs for lossless object exchange at the runtime boundary; \textit{(ii)} an Agent-Skills--compatible\footnote{Agent Skills is an open specification for portable skill packaging \citep{anthropic2025agentskills}; see Section~\ref{sec:skills} for details of CaveAgent's extension.} packaging layer that delivers executable extensions --- not merely text instructions --- into the runtime (Section~\ref{sec:skills}); and \textit{(iii)} runtime-mediated multi-agent coordination demonstrated qualitatively in Appendix~\ref{app_multiagent} (rigorous quantitative evaluation is future work). MemGPT \citep{packer2023memgpt} addresses long-context memory via tiered text summarization --- an orthogonal axis to runtime-state management. TaskWeaver \citep{qiao2023taskweaver} preserves data structures across turns and provides a framework-specific plugin format that predates the Agent Skills standard; inter-agent state sharing is also not part of its design. Among the four frameworks, only CaveAgent provides first-class bidirectional object exchange together with Agent Skills compatibility, the architectural combination that underpins the experimental gains reported in Section~\ref{exp}.

\begin{table}[ht]
\centering
\caption{Capability comparison: CaveAgent vs.\ closely-related stateful agent frameworks. \cmark: supported as a first-class capability; ``partial'': supported with caveats (see footnotes); \xmark: not part of the system's design. CaveAgent's column is shaded for emphasis.}
\label{tab:framework_comparison}
\small
\renewcommand{\arraystretch}{1.2}
\begin{tabular}{@{}p{0.32\textwidth}cccc@{}}
\toprule
\textbf{Capability} & \cellcolor{highlightcolor}\textbf{CaveAgent} & \textbf{CodeAct} & \textbf{MemGPT} & \textbf{TaskWeaver} \\
\midrule
Persistent code-execution kernel & \cellcolor{highlightcolor}\cmark & \cmark & \xmark & \cmark \\
External\,$\rightarrow$\,runtime variable injection & \cellcolor{highlightcolor}\cmark & \xmark & \xmark & partial$^\dagger$ \\
Runtime\,$\rightarrow$\,external object retrieval & \cellcolor{highlightcolor}\cmark & \xmark & \xmark & partial$^\dagger$ \\
Lossless object exchange at runtime boundary & \cellcolor{highlightcolor}\cmark & \xmark & \xmark & \xmark \\
Agent Skills open-standard support & \cellcolor{highlightcolor}\cmark & \xmark & \xmark & \xmark$^\ddagger$ \\
Runtime-mediated multi-agent coordination & \cellcolor{highlightcolor}\cmark$^\S$ & \xmark & \xmark & \xmark \\
\bottomrule
\end{tabular}

\vspace{0.5em}
\footnotesize
$^\dagger$\,TaskWeaver exposes tools and data to its runtime via plugins, but lacks a first-class API for injecting or retrieving arbitrary external Python objects directly.\quad
$^\ddagger$\,TaskWeaver (Nov.\ 2023) predates the Agent Skills standard (Dec.\ 2025); its plugin format is framework-specific.\quad
$^\S$\,Demonstrated qualitatively in Appendix~\ref{app_multiagent}; rigorous quantitative evaluation is left for future work.
\end{table}


\section{CaveAgent: Stateful Runtime Management}
\subsection{Core Methodologies}
\begin{figure}
\centering
\includegraphics[width=\textwidth]{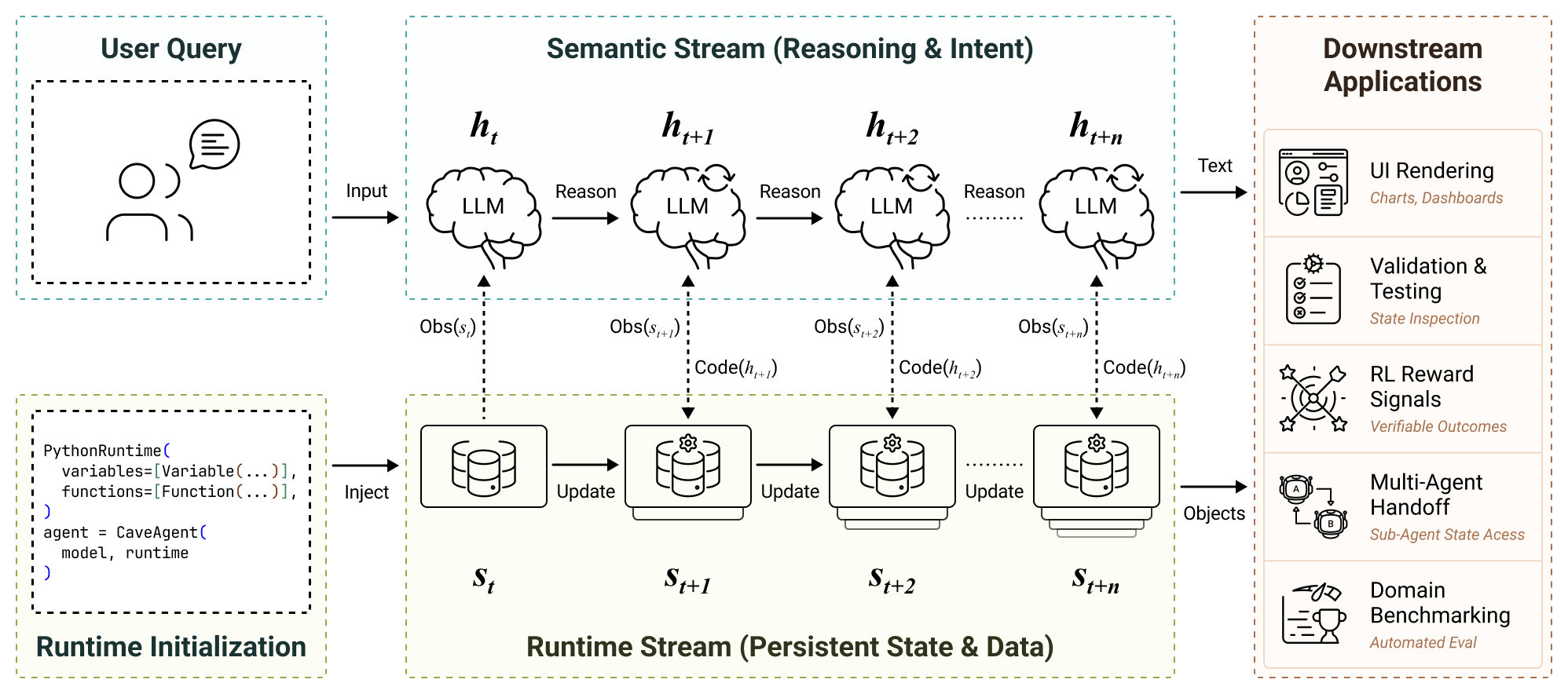}
\caption{Framework Overview}
\label{framework}
\end{figure}

CaveAgent adopts a \textbf{dual-stream architecture} (Figure~\ref{framework}): a \textit{Semantic Stream} for lightweight reasoning, and a \textit{Runtime Stream} for stateful execution. Unlike conventional agents where the LLM's text context serves as the primary workspace and tools are auxiliary services, CaveAgent inverts this relationship: the persistent runtime becomes the central locus of data storage, computation, and state management, while the semantic stream is reduced to a lightweight controller generating code to operate on the runtime.

We model the agent's task as a sequential decision process over a horizon $T$. At each turn $t \in [1, T]$, the agent receives a query or observation $x_t$ and must produce a response $y_t$. Unlike traditional formulations where the entire state is re-serialized into $x_t$, we introduce a latent runtime state $\mathcal{S}_t$ (we call it "in-runtime context"). The system evolution is thus defined by:
\begin{align}
    h_t &= \text{LLM}(x_t, h_{t-1}) \quad &(\text{Semantic Stream: Context History}) \\
    \mathcal{S}_t &= \text{Exec}(c_t, \mathcal{S}_{t-1}) \quad &(\text{Runtime Stream: Persistent Environment})
\end{align}
where $h_t$ represents the semantic history (``in-prompt context'') and $c_t$ is the executable code generated by the agent. The key design choice is the decoupling of $h_t$ and $\mathcal{S}_t$: the semantic stream tracks \textit{intent} and lightweight \textit{reasoning} for code generation, while the runtime stream maintains all \textit{data} and \textit{execution state} via the code generated by the semantic stream.

\paragraph{The Runtime Stream:}
The execution kernel of the runtime stream is a persistent Python kernel (an IPython interactive shell). We conceptualize each interaction turn $t$ not as an isolated API call, but as a \textbf{cell execution} in a virtual Jupyter notebook.
\begin{itemize}
    \item \textbf{Persistent Namespace:} The state $\mathcal{S}_t$ comprises the global namespace $\mathcal{N}_t$, containing all variables, functions, and imported modules. When the agent executes code $c_t$ (e.g., \texttt{x = 5}), the modification to $\mathcal{N}_t$ persists to $\mathcal{N}_{t+1}$. This allows subsequent turns to reference \texttt{x} directly without requiring the LLM to memorize or re-output its value.
    \item \textbf{Stateful Injection:} Tools are not only described in text; they are \textit{injected} into $\mathcal{N}_0$ as live Python objects. This allows the agent to interact with stateful objects via calling tools that modify the object's internal state across turns.
\end{itemize}
The runtime stream can also assign values to new variables during interaction and inject them into the persistent namespace (in-runtime context). This enables large context in complex tasks, such as large DataFrames, graphs, or other data structures, to be managed entirely by the Python runtime stream as stateful variables. Their values are preserved natively in persistent runtime memory without requiring repeated serialization into text, eliminating the risk of hallucination from lossy textual representations.

\textbf{Code as Action, State as Memory} \quad The agent stores key information (such as reasoning chains and intermediate data analysis results) as persistent variables in the runtime context, retaining only a lightweight description and reference in its in-prompt context. The runtime thus functions as an external memory, allowing the agent to retrieve stored data as native Python objects, achieving context compression and reducing progressive context degradation in multi-turn interactions. This property addresses persistent challenges in agentic tool use, specifically memory, dynamic decision-making, and long-horizon reasoning \citep{bfcl}.

A natural question arises: what distinguishes storing intermediate results in runtime variables from persisting them to files? We identify three key differences. First, \textit{type fidelity}: runtime variables preserve native Python object types with full method interfaces (e.g., a DataFrame retains \texttt{.groupby()}, \texttt{.merge()} operations), whereas file-based storage requires serialization that may lose type information or fail entirely for non-serializable objects such as database connections, trained models with custom layers, or open file handles. Second, \textit{access latency}: runtime variables enable zero-cost retrieval within the same memory space, while file I/O introduces disk latency and requires explicit read/write operations in generated code. Third, \textit{lifecycle management}: runtime variables are automatically scoped to the agent session and garbage-collected appropriately, whereas file-based approaches require explicit cleanup logic to avoid accumulating temporary artifacts. That said, file-based persistence offers advantages for large-scale data exceeding memory capacity and for checkpointing across agent restarts. We note that CaveAgent also supports storing intermediate results in files, but storing them in runtime variables better respects the persistent runtime paradigm, yielding a more unified system.

Programmatic state retrieval enables the extraction of manipulated Python objects for direct use in downstream applications. Unlike conventional agents that produce text outputs requiring parsing and reconstruction, CaveAgent exposes native objects (DataFrames, class instances, arrays) with full type fidelity. This enables UI rendering via direct object binding, RL reward computation through programmatic state inspection, validation via unit test assertions against returned structures, and lossless object passing in multi-agent systems (Appendix \ref{app_multiagent}). The agent thus transforms the LLM from an isolated text generator into the operator of a stateful, interoperable computational component.

\paragraph{The Semantic Stream:}
Parallel to the runtime stream, the semantic stream uses the LLM to generate code that manipulates the runtime. It is also responsible for:
\begin{itemize}
    \item \textbf{Prompt Construction:} Dynamically generating system instructions that describe the \textit{signatures} of available tools in $\mathcal{N}_t$, without dumping their full state (which may be large) into the in-prompt context window.
    \item \textbf{Observation Shaping:} Captures execution outputs and enforces a length constraint $\tau(\cdot)$ to prevent context explosion. This feedback mechanism guides the agent to interact with the persistent state efficiently, prioritizing concise and relevant information over verbose raw dumps in the in-prompt context $h_{t+1}$.
\end{itemize}

This split addresses the ``Context Explosion'' problem: large data remains in $\mathcal{S}_t$ while only high-level reasoning flows through $h_t$, avoiding the token overhead inherent in text-centric architectures. The bidirectional interface for \textbf{injecting} and \textbf{retrieving} structured objects of any type distinguishes CaveAgent from JSON-based function calling \citep{toolsandbox} and from code-based approaches with internalized runtimes; detailed contrasts appear in Section~\ref{sec:framework_comparison}. Algorithm~\ref{alg:cave_agent} (Appendix~\ref{algo_appendix}) shows the iteration loop; we next describe CaveAgent's core mechanisms.

\subsubsection{Variable and Function Injection}
CaveAgent treats Python objects and functions as first-class citizens within the runtime. Each injectable entity is wrapped in a container that automatically extracts metadata: signatures, type hints, and docstrings for functions; names, types, and descriptions for variables. This metadata is aggregated into the system prompt as a lightweight ``API reference,'' while the actual objects are mapped directly into the execution engine's namespace as global symbols. This design enables \textit{Object-Oriented Interaction}: instead of stateless JSON calls (e.g., \texttt{tool: "sort", args: \{...\}}), the model invokes methods on stateful objects directly (e.g., \texttt{processor.process(data)}), chaining method calls and manipulating attributes naturally. The agent interacts via executable Python programs with native control flow (loops, conditionals) and stateful data passing, delivering final output as a native Python object rather than a textual approximation.

\subsubsection{Dynamic Context Synchronization}
The Semantic Stream is ``blind'' to the Runtime Stream by default. To inspect runtime state, the agent must explicitly generate code (e.g., \texttt{print(df.head())}), enforcing an \textbf{Active Attention} mechanism that selectively pulls only relevant slices into the token context. To prevent context explosion from verbose outputs, an \textbf{Observation Shaping} layer enforces a length constraint: when output exceeds $L_{\max}$, the system returns a structured error prompting the agent to use summary methods. This feedback loop guides efficient interaction with persistent state.

\subsubsection{Security Check via Static Analysis}
CaveAgent mitigates code execution risks via AST-based static analysis with modular policy rules: \textit{ImportRule} (blocking unauthorized modules), \textit{FunctionRule} (prohibiting dangerous calls like \texttt{eval()}), and \textit{AttributeRule} (preventing sandbox bypass). Violations return structured errors to the semantic stream, enabling self-correction without breaking interaction continuity.

\subsection{Runtime-Integrated Skill Management}
\label{sec:skills}

\begin{figure}[t]
    \centering
    \includegraphics[width=0.55\linewidth]{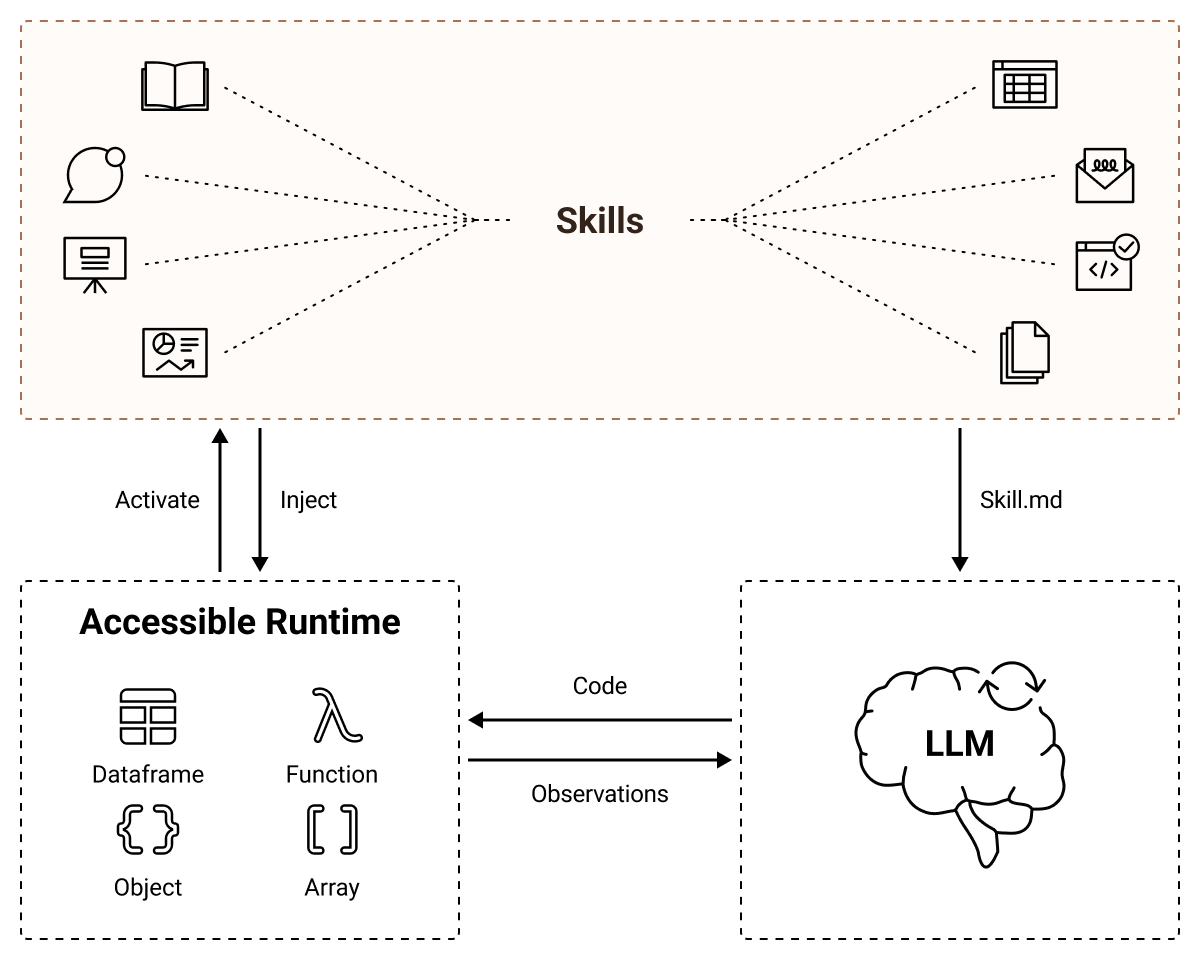}
    \caption{CaveAgent's Skill Management extends the Agent Skills open standard with runtime injection. Standard skills provide text prompts; CaveAgent skills additionally export functions, variables, and types into the persistent runtime.}
    \label{fig:skills}
\end{figure}

CaveAgent extends the Agent Skills open standard (see Section~\ref{sec:framework_comparison}) by introducing an \texttt{injection.py} module alongside the standard \texttt{SKILL.md} file. While standard skills provide text-based prompts that guide LLM behavior, CaveAgent skills additionally export \textbf{Functions}, \textbf{Variables}, and \textbf{Type} definitions that are injected directly into the persistent runtime upon activation. The framework employs progressive disclosure: skill metadata (name and description) is loaded at startup for routing decisions, while full instructions and runtime injections are loaded on-demand when the agent invokes \texttt{activate\_skill()}. This bridges declarative skill definitions with CaveAgent's stateful runtime paradigm: skills deliver executable artifacts into the runtime, not merely textual instructions to the LLM. As shown in Figure~\ref{fig:skills}, domain expertise is packaged as both human-readable instructions for the language model and machine-executable artifacts for the runtime.

\begin{table}[t]
\centering
\caption{Comparison of Standard Agent Skills vs. CaveAgent Skills.}
\label{tab:skills_comparison}
\small
\begin{tabular}{@{}lcc@{}}
\toprule
\textbf{Property} & \textbf{Standard Skills} & \textbf{CaveAgent Skills} \\
\midrule
Skill format & \texttt{SKILL.md} & \texttt{SKILL.md} + \texttt{injection.py} \\
LLM interaction & Reads text prompts & Operates on injected objects \\
Capability delivery & Textual instructions & Executable runtime artifacts \\
Data flow & Text-in, text-out & Object-in, object-out \\
Tool delivery & Cannot deliver tools & Injects tools into runtime \\
\bottomrule
\end{tabular}
\end{table}

\paragraph{Discussion: From Tool Disclosure to Object Disclosure.}
The skill activation pattern (lazy, metadata-driven retrieval of executable artifacts) generalizes beyond tools to any runtime object. Intermediate results stored as persistent variables (``Code as Action, State as Memory'') can similarly be discovered via semantic search over their metadata (names, types, descriptions), bringing relevant state into the agent's context only when needed. Unlike retrieval-augmented generation over text chunks, the retrieved items are live Python objects with full type fidelity. This extends progressive disclosure from tool management to general workflow-state management.


\section{Experiments}
\label{exp}

In this section, we validate CaveAgent by answering four questions:
\begin{itemize}
    \item \textbf{[Q1.]}  Can CaveAgent perform on par with or surpass standard function-calling paradigms on widely-used benchmarks involving \textbf{basic} function-calling tasks? This is to showcase the basic function calling capabilities of CaveAgent.
    \item \textbf{[Q2.]} Can CaveAgent successfully perform state management across multi-turns correctly and efficiently?
    \item \textbf{[Q3.]} How token-efficient is CaveAgent compared to traditional JSON-based and Codeact style function calling?
    \item \textbf{[Q4.]} How does CaveAgent adapt to complex scenarios that require manipulating complex data objects? This is to showcase CaveAgent's unique advantages.
\end{itemize}

\subsection{[\textbf{Q1}] Standard Function Calling Benchmarks}
We evaluate on Tau$^2$-bench~\citep{barres2025tau2} and BFCL~\citep{bfcl} using six SOTA LLMs: \textbf{DeepSeek-V3.2} (685B MoE), \textbf{Qwen3 Coder} (30B MoE), \textbf{Kimi K2 0905} (1000B MoE), \textbf{Claude Sonnet 4.5}, \textbf{GPT-5.1}, and \textbf{Gemini 3 Pro}. For each model, we compare its native function-calling mechanism against CaveAgent, where the LLM serves solely as a text generation engine bypassing internal function-calling modules. Per-model sampling configurations are listed in Table~\ref{tab:model_setup}.

\begin{table}[!htbp]
\centering
\caption{Model details and sampling configuration for the Tau$^2$-bench evaluation. Architecture column reports MoE configuration as \textit{total (active) parameters}; ``N/A'' indicates that the parameter count or architecture has not been publicly disclosed.}
\label{tab:model_setup}
\small
\setlength{\tabcolsep}{4pt}
\renewcommand{\arraystretch}{1.15}
\begin{tabular}{@{}llllll@{}}
\toprule
\textbf{Model} & \textbf{Reasoning} & \textbf{Size} & \textbf{Temp.} & \textbf{Arch.} & \textbf{Temperature notes} \\
\midrule
Qwen3-Coder 30B       & None          & 30B (3B active)    & 0.2 & MoE & For stable code generation \\
Kimi-K2-0905          & None          & 1000B (32B active) & 0.6 & MoE & Official recommendation \\
DeepSeek-V3.2         & None          & 685B (37B active)  & 0.2 & MoE & For stable code generation \\
Claude Sonnet 4.5     & None          & N/A                & 0.2 & N/A & For stable code generation \\
GPT-5.1               & None          & N/A                & 1.0 & N/A & Only default value supported \\
Gemini 3 Pro Preview  & Low thinking  & N/A                & 1.0 & N/A & Official recommendation \\
\bottomrule
\end{tabular}
\end{table}

\subsubsection{Results on Tau$^2$-bench}
Tau$^2$-bench evaluates multi-turn tool use in realistic conversational scenarios (Airline and Retail domains), requiring agents to maintain consistency across turns. We follow the original evaluation protocols with DeepSeek V3 as user simulator, testing each model three times per domain. Since CaveAgent executes Python code, we employ runtime instrumentation with wrapper functions to capture and compare function invocations against ground truth, ensuring fair cross-paradigm evaluation.

\begin{table}[t]
\centering
\caption{Performance comparison on the Tau$^2$ Benchmark across different models and domains ($n{=}3$ runs). Avg.\ values are reported as $\text{mean}{\pm}\text{std}$ where std is the sample (Bessel-corrected) standard deviation across the three runs. \textbf{Bold} indicates that CaveAgent's mean exceeds Function Calling's mean; see the variance analysis paragraph below for which gains exceed run-to-run variance. Cell values are task success rates (\%).}
\label{tab:Tau$^2$_bench}
\small
\setlength{\tabcolsep}{3.5pt}
\renewcommand{\arraystretch}{1.1}
\begin{tabular}{@{}ll ccc g ccc g@{}}
\toprule
\multirow{2}{*}{\textbf{Model}} & \multirow{2}{*}{\textbf{Domain}} & \multicolumn{4}{c}{\textbf{Function Calling}} & \multicolumn{4}{c}{\textbf{CaveAgent}} \\
\cmidrule(lr){3-6} \cmidrule(lr){7-10}
 &  & Run 1 & Run 2 & Run 3 & \cellcolor{avgcolor}Avg.\textuparrow & Run 1 & Run 2 & Run 3 & \cellcolor{avgcolor}Avg.\textuparrow \\
\midrule
\multicolumn{10}{l}{\textbf{\textit{Open Source}}} \\
\midrule
\multirow{2}{*}{\shortstack[l]{DeepSeek-V3.2\\{\scriptsize (685B)}}}
 & Airline & 56.0 & 56.0 & 54.0 & 55.3{\scriptsize$\pm$1.2} & 62.0 & 60.0 & 58.0 & \textbf{60.0}{\scriptsize$\pm$2.0} {\scriptsize \textcolor{teal}{(+4.7)}} \\
 & Retail  & 79.8 & 77.2 & 74.6 & 77.2{\scriptsize$\pm$2.6} & 85.1 & 82.5 & 78.1 & \textbf{81.9}{\scriptsize$\pm$3.5} {\scriptsize \textcolor{teal}{(+4.7)}} \\
\midrule
\multirow{2}{*}{\shortstack[l]{Qwen3-Coder\\{\scriptsize (30B)}}}
 & Airline & 36.0 & 40.0 & 38.0 & 38.0{\scriptsize$\pm$2.0} & 36.0 & 42.0 & 44.0 & \textbf{40.7}{\scriptsize$\pm$4.2} {\scriptsize \textcolor{teal}{(+2.7)}} \\
 & Retail  & 41.2 & 43.0 & 39.5 & 41.2{\scriptsize$\pm$1.8} & 51.8 & 54.4 & 57.9 & \textbf{54.7}{\scriptsize$\pm$3.1} {\scriptsize \textcolor{teal}{(+13.5)}} \\
\midrule
\multirow{2}{*}{\shortstack[l]{Kimi-K2-0905\\{\scriptsize (1000B)}}}
 & Airline & 52.0 & 56.0 & 54.0 & 54.0{\scriptsize$\pm$2.0} & 58.0 & 54.0 & 54.0 & \textbf{55.3}{\scriptsize$\pm$2.3} {\scriptsize \textcolor{teal}{(+1.3)}} \\
 & Retail  & 62.3 & 60.5 & 59.6 & 60.8{\scriptsize$\pm$1.4} & 69.3 & 72.8 & 71.9 & \textbf{71.3}{\scriptsize$\pm$1.8} {\scriptsize \textcolor{teal}{(+10.5)}} \\
\midrule
\multicolumn{10}{l}{\textbf{\textit{Closed Source}}} \\
\midrule
\multirow{2}{*}{Claude Sonnet 4.5}
 & Airline & 56.0 & 54.0 & 62.0 & 57.3{\scriptsize$\pm$4.2} & 56.0 & 52.0 & 62.0 & 56.7{\scriptsize$\pm$5.0} {\scriptsize \textcolor{gray}{(-0.7)}} \\
 & Retail  & 68.4 & 67.5 & 81.6 & 72.5{\scriptsize$\pm$7.9} & 73.7 & 75.4 & 80.7 & \textbf{76.6}{\scriptsize$\pm$3.7} {\scriptsize \textcolor{teal}{(+4.1)}} \\
\midrule
\multirow{2}{*}{GPT-5.1}
 & Airline & 50.0 & 58.0 & 50.0 & 52.7{\scriptsize$\pm$4.6} & 58.0 & 56.0 & 54.0 & \textbf{56.0}{\scriptsize$\pm$2.0} {\scriptsize \textcolor{teal}{(+3.3)}} \\
 & Retail  & 64.0 & 66.7 & 66.7 & 65.8{\scriptsize$\pm$1.6} & 65.8 & 69.3 & 73.6 & \textbf{69.6}{\scriptsize$\pm$3.9} {\scriptsize \textcolor{teal}{(+3.8)}} \\
\midrule
\multirow{2}{*}{Gemini 3 Pro}
 & Airline & 64.0 & 62.0 & 58.0 & 61.3{\scriptsize$\pm$3.1} & 68.0 & 68.0 & 68.0 & \textbf{68.0}{\scriptsize$\pm$0.0} {\scriptsize \textcolor{teal}{(+6.7)}} \\
 & Retail  & 72.8 & 72.8 & 66.7 & 70.8{\scriptsize$\pm$3.5} & 77.2 & 76.3 & 75.4 & \textbf{76.3}{\scriptsize$\pm$0.9} {\scriptsize \textcolor{teal}{(+5.5)}} \\
\bottomrule
\end{tabular}
\end{table}

\paragraph{Performance Analysis.}
The results on Tau$^2$-bench are summarized in Table~\ref{tab:Tau$^2$_bench}. Key findings include:

\textit{(1).} CaveAgent outperforms JSON-based function calling in 11 out of 12 settings across models from 30B to over 1000B parameters, with consistent improvements for \textbf{DeepSeek-V3.2} (\textbf{+4.7\%}) and \textbf{Gemini 3 Pro} (\textbf{+6.1\%}), showing that offloading state management to a deterministic code runtime improves performance.

\textit{(2).} Gains are amplified in state-intensive \textit{Retail} scenarios, where complex transaction modifications require maintaining state consistency across turns. CaveAgent achieves double-digit gains for Qwen3 and Kimi K2, validating that \textbf{Stateful Runtime Management} reduces serialization-induced errors (detailed trajectory analysis in Appendix \ref{Additional_case:tau2}).

\textit{(3).} The code-specialized \textbf{Qwen3-Coder (30B)} exhibits the largest improvement (\textbf{+13.5\%} in Retail), rivaling larger models. CaveAgent \textbf{leverages the inherent coding proficiency of LLMs}, allowing code-centric models to focus on logic generation rather than verbose context tracking.

\textit{(4) Variance analysis.} With $n{=}3$ runs we report sample standard deviations alongside the per-condition means in Table~\ref{tab:Tau$^2$_bench}; the combined standard deviation for a difference of means is $\sigma_{\text{c}} = \sqrt{(\sigma_{\text{FC}}^2 + \sigma_{\text{CA}}^2)/n}$. Several Airline-domain gains lie within or close to run-to-run noise: Claude Sonnet 4.5's small Airline regression ($-0.7$, $-0.2\sigma_{\text{c}}$) is statistically indistinguishable from zero; Kimi-K2 ($+1.3$, $0.8\sigma_{\text{c}}$) and Qwen3-Coder ($+2.7$, $1.0\sigma_{\text{c}}$) Airline gains lie at or within $1\sigma_{\text{c}}$; and GPT-5.1's Airline gain ($+3.3$, $1.2\sigma_{\text{c}}$) is only marginally above. By contrast, Airline gains for \textbf{DeepSeek-V3.2} ($+4.7$, $3.5\sigma_{\text{c}}$) and \textbf{Gemini 3 Pro} ($+6.7$, $3.8\sigma_{\text{c}}$) clearly exceed $2\sigma_{\text{c}}$. On the Retail domain, gains are markedly more robust overall: \textbf{Qwen3-Coder} ($+13.5$, $6.6\sigma_{\text{c}}$), \textbf{Kimi-K2} ($+10.5$, $8.0\sigma_{\text{c}}$), \textbf{Gemini~3 Pro} ($+5.5$, $2.6\sigma_{\text{c}}$), \textbf{DeepSeek-V3.2} ($+4.7$, $1.9\sigma_{\text{c}}$), and \textbf{GPT-5.1} ($+3.8$, $1.6\sigma_{\text{c}}$) all exceed $1.5\sigma_{\text{c}}$; the one exception is \textbf{Claude Sonnet 4.5} ($+4.1$, $0.8\sigma_{\text{c}}$), where a high Function-Calling baseline variance ($\sigma_{\text{FC}}{=}7.9$, driven by an outlying run-3 score of 81.6 against 67.5 and 68.4) absorbs the gain. The pattern aligns with our overall thesis: CaveAgent's stateful runtime management offers its most robust advantage on multi-turn, state-intensive workflows (Retail), where serialization overhead and context drift accumulate; on the lighter-weight Airline domain, the architectural advantage is smaller and run-to-run variance can dominate at $n{=}3$. We acknowledge this limitation and note that larger-$n$ replications, beyond the API-cost budget of this study, would tighten the per-condition confidence intervals.

\subsubsection{Results on BFCL}
To complement Tau$^2$-bench's multi-turn evaluation, we assess atomic function-calling precision on the \textbf{Berkeley Function Calling Leaderboard (BFCL) v3}~\citep{bfcl}. We evaluate on the four expert-curated single-turn categories of BFCL v3 --- \texttt{simple} (400 entries), \texttt{multiple} (200), \texttt{parallel} (200), and \texttt{parallel\_multiple} (200), totaling 1{,}000 question-function-answer pairs of increasing structural complexity. Because BFCL v3 also includes \textit{live}, \textit{multi-turn}, and \textit{multi-step} subsets, we restrict evaluation to the four AST-evaluated categories to keep this benchmark disjoint from our multi-turn evaluation on Tau$^2$-bench. We use Executable Evaluation (functional correctness) by executing generated code and comparing results against ground truth. The summary is shown in Table~\ref{tab:bfcl_summary} (detailed per-run results in Appendix Table~\ref{tab:bfcl}).

\begin{table}[t]
\centering
\caption{BFCL Benchmark Summary (avg. of 3 runs). \textbf{Bold} indicates CaveAgent outperforms Function Calling.}
\label{tab:bfcl_summary}
\small
\begin{tabular}{@{}lccc@{}}
\toprule
\textbf{Model} & \textbf{FC Avg.(\%)} & \textbf{CaveAgent Avg.(\%)} & \textbf{$\Delta$} \\
\midrule
DeepSeek-V3.2 (685B) & 86.9 & \textbf{94.0} & \textcolor{teal}{+7.1} \\
DeepSeek-V3.2 (w/o prompt) & 53.1 & \textbf{94.0} & \textcolor{teal}{+40.9} \\
Qwen3-Coder (30B) & 89.8 & \textbf{94.4} & \textcolor{teal}{+4.6} \\
Kimi-K2-0905 (1000B) & 89.2 & \textbf{94.7} & \textcolor{teal}{+5.5} \\
Claude Sonnet 4.5 & 94.4 & 94.4 & \textcolor{gray}{~0.0} \\
GPT-5.1 & 89.6 & 88.9 & \textcolor{gray}{-0.7} \\
Gemini 3 Pro & 94.3 & 94.3 & \textcolor{gray}{~0.0} \\
\bottomrule
\end{tabular}
\end{table}

\paragraph{Performance Analysis.}
DeepSeek-V3.2 without explicit parallel-execution prompting achieves only 53.1\% under the JSON paradigm due to its strong inductive bias toward sequential execution \citep{liu2025deepseek}. CaveAgent achieves \textbf{94.0\%} without any prompt intervention, as Python code naturally supports parallel execution via independent statements while preserving inter-tool dependency reasoning. The 30B \textbf{Qwen3-Coder} with CaveAgent (94.4\%) outperforms the much larger \textbf{GPT-5.1} (89.6\%) and matches \textbf{Claude Sonnet 4.5}, demonstrating that CaveAgent leverages the coding proficiency of smaller LLMs. For SOTA models already at near-ceiling performance (Claude Sonnet 4.5, Gemini 3 Pro), gains are negligible since remaining errors stem from ambiguous queries rather than model incapacity. The advantages of our paradigm are most apparent in tasks requiring manipulation of \textbf{complex data objects} over long-horizon interactions, which we assess next.

\subsection{[\textbf{Q2}] Case Study: Stateful Management}
We design a benchmark targeting dimensions of state manipulation that existing benchmarks do not address, measuring an agent's ability to read, modify, and persist variables across turns. A key design principle is \textit{programmatic validation}: we directly inspect runtime state after execution against ground-truth expectations, rather than parsing text outputs. For each dimension, we curate test cases with linearly dependent queries and initial variable states (see Appendix \ref{cases}). Results are shown in Table \ref{statebench}.

\begin{table}[t]
    \centering
    \caption{Stateful Management Benchmark Results (success rate \%) across three evaluation dimensions. (1) For Type Proficiency, we designed 36, 36, and 42 cases for Simple, Object, and Scientific types, respectively. (2) For Multi-variable Stateful Management, we established 15 evaluation points for each variable-count tier. (3) For Multi-turn Stateful Management, we developed two scenarios, each consisting of 40 turns distributed across two conversations. }
    \resizebox{\textwidth}{!}{%
        \begin{tabular}{l ccc c ccccc c cc c}
            \toprule
            & \multicolumn{4}{c}{\textbf{Type Proficiency (\%)}} & \multicolumn{6}{c}{\textbf{Multi-Variable (\%)}} & \multicolumn{3}{c}{\textbf{Multi-Turn (\%)}} \\
            \cmidrule(lr){2-5} \cmidrule(lr){6-11} \cmidrule(lr){12-14}
            \textbf{Model} & \makecell{Simple\\(36)} & \makecell{Object\\(36)} & \makecell{Sci.\\(42)} & \textbf{Avg} & \makecell{5V\\(15)} & \makecell{10V\\(15)} & \makecell{15V\\(15)} & \makecell{20V\\(15)} & \makecell{25V\\(15)} & \textbf{Avg} & \makecell{Home\\(40)} & \makecell{Fin.\\(40)} & \textbf{Avg} \\
            \midrule
            \cellcolor{highlightcolor}\textbf{DeepSeek-V3.2} & \cellcolor{highlightcolor}\textbf{100} & \cellcolor{highlightcolor}\textbf{100} & \cellcolor{highlightcolor}\textbf{100} & \cellcolor{highlightcolor}\textbf{100} & \cellcolor{highlightcolor}\textbf{100} & \cellcolor{highlightcolor}\textbf{100} & \cellcolor{highlightcolor}\textbf{100} & \cellcolor{highlightcolor}\textbf{100} & \cellcolor{highlightcolor}\textbf{100} & \cellcolor{highlightcolor}\textbf{100} & \cellcolor{highlightcolor}\textbf{100} & \cellcolor{highlightcolor}\textbf{100} & \cellcolor{highlightcolor}\textbf{100} \\
            Qwen3 Coder & 100 & 94.4 & 95.2 & 96.5 & 94.4 & 100 & 80.0 & 80.0 & 100 & 90.9 & 77.5 & 85.0 & 81.3 \\
            Kimi K2 0905 & 100 & 100 & 100 & 100 & 100 & 100 & 100 & 100 & 100 & 100 & 90.0 & 100 & 95.0 \\
            Gemini 3 Pro & 100 & 100 & 100 & 100 & 100 & 100 & 100 & 100 & 100 & 100 & 97.5 & 100 & 98.7 \\
            \bottomrule
        \end{tabular}%
    }
    \label{statebench}
\end{table}

\paragraph{Type Proficiency} evaluates manipulation of Python primitives, user-defined class instances, and scientific types (DataFrames, ndarrays). Results yield uniformly high scores (96.5\%--100\%), validating that code-based manipulation of complex types is tractable for current LLMs.

\paragraph{Multi-Variable} tests how accuracy scales with 5--25 concurrent variables across five tiers (15 evaluation points each). Top models maintain 100\% accuracy throughout, demonstrating that concurrent state management scales effectively within CaveAgent's architecture.

\paragraph{Multi-Turn} assesses state persistence across 40-turn interactions in two scenarios: Smart Home (device state consistency) and Financial Account (numerical precision over multi-step operations). While DeepSeek-V3.2 maintains perfect accuracy, other models exhibit degradation on long-horizon state tracking. The consistently high accuracy across top models validates our thesis: when LLMs interact through code with persistent runtime state, \textbf{reliable} and \textbf{verifiable} agent behavior becomes achievable.

\paragraph{Discriminability note.} Type Proficiency ($96.5$--$100\%$) and Multi-Variable ($90.9$--$100\%$) compress most evaluated models near the ceiling, providing limited model-ranking signal. We retain both dimensions because their purpose is \textit{structural} rather than competitive: a near-100\% score validates that CaveAgent's runtime architecture \textit{does} reliably support manipulation of any tested object type and any concurrent variable count up to 25, removing a class of potential architectural failure modes from subsequent claims. Multi-Turn ($81.3$--$98.7\%$ across the non-saturated models) is the most discriminative dimension and serves as the model-ranking signal of this section, stressing the 40-turn state persistence regime where in-context tracking would otherwise accumulate drift. Designing strictly harder Type-Proficiency and Multi-Variable tasks (e.g., adversarial type coercion across boundaries, $50+$ concurrent variables under interleaved updates) is a natural extension and is left for future work.

\begin{figure}[t]
    \centering
    \begin{subfigure}[b]{0.49\textwidth}
        \centering
        \includegraphics[width=\textwidth]{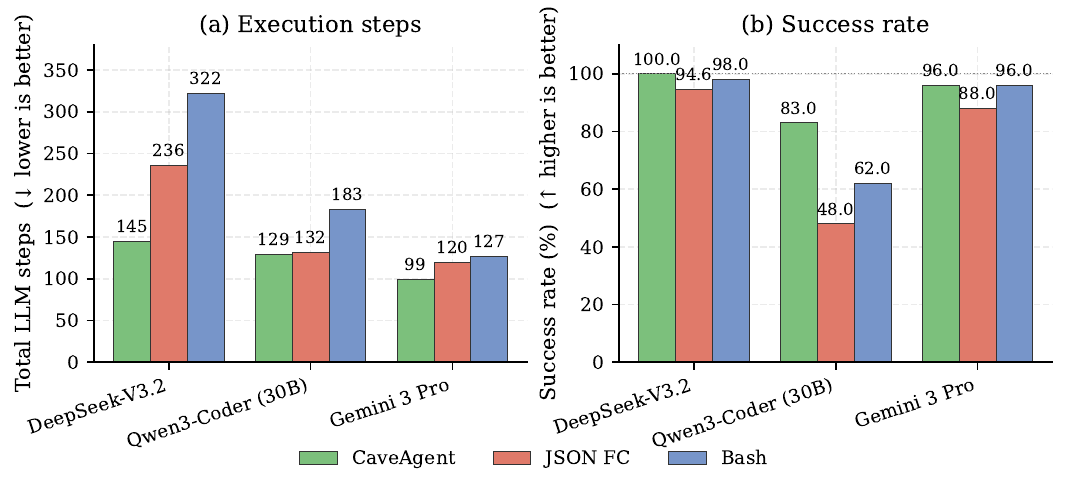}
        \caption{Compare Execution steps and task success rates.}
        \label{token:left}
    \end{subfigure}
    \hfill
    \begin{subfigure}[b]{0.49\textwidth}
        \centering
        \includegraphics[width=\textwidth]{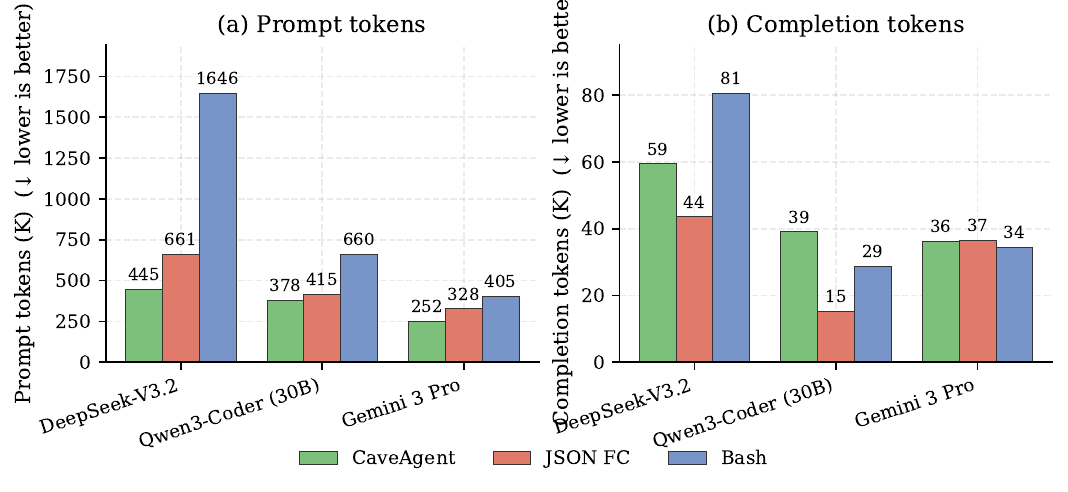}
        \caption{Compare Prompt tokens and completion tokens.}
        \label{token:right}
    \end{subfigure}

    \caption{Q3 token-efficiency comparison across three models (DeepSeek-V3.2, Qwen3-Coder 30B, Gemini~3~Pro) and three paradigms (CaveAgent, JSON-based function calling, and a bash filesystem agent), focusing on execution steps, success rates, and prompt / completion token consumptions, aggregated over three scenarios (IoT, finance, e-commerce). Here, the steps means the number of turns needed for task completion, prompt tokens refers to the cumulative input tokens sent to the model across all turns (including system prompts, conversation history, and tool results), and completion tokens refers to the cumulative output tokens generated by the model (including reasoning, function calls, or code generation). Aggregate numerical values are also reported in Table~\ref{tab:Q3_aggregate}.}
    \label{token}
\end{figure}

\subsection{[\textbf{Q3}] Token Efficiency Study}
\paragraph{Setup.}
We evaluate token efficiency across three domains (IoT, finance, e-commerce) with logically interdependent tool operations. We extend the original evaluation along two axes: (i)~two additional model families, Qwen3-Coder (30B) and Gemini~3~Pro,\footnote{We use \texttt{qwen3-coder-30b-a3b-instruct} (the canonical 30B-class identifier) and \texttt{gemini-3.1-pro-preview} as a drop-in substitute for the deprecated \texttt{gemini-3-pro-preview}.} matched to the model labels of Section~\ref{exp}; and (ii)~a third paradigm, a single-tool bash agent that persists state through the filesystem. The bash agent exposes a single tool \texttt{bash(command: str)} with a per-conversation sandbox; scenario tools are reached through \texttt{python -c "import tools; ..."}, and the system prompt explicitly directs filesystem persistence with worked dump\,$\rightarrow$\,load examples. All runs share the same scenarios, validators, and \texttt{max\_steps}~$=40$; sampling temperatures follow the per-model defaults of Table~\ref{tab:model_setup} --- $1.0$ for Gemini~3~Pro (its only supported value) and $0.2$ for DeepSeek-V3.2 and Qwen3-Coder (for stable code generation) --- with $n{=}1$ per cell. Aggregate results across the three domains are reported in Table~\ref{tab:Q3_aggregate} and visualised in Figure~\ref{token}.

\begin{table}[t]
\centering
\caption{Q3 Token Efficiency Study: aggregate results across three domains (IoT, finance, e-commerce) for each (model, paradigm) cell. The two DeepSeek-V3.2 rows for CaveAgent and JSON FC reproduce the original Q3 results; the remaining rows extend the study to two additional model families and to a third paradigm (a bash filesystem agent). Best success rate per model is shown in bold; CaveAgent rows are shaded for readability. $n{=}1$ per cell.}
\label{tab:Q3_aggregate}
\small
\setlength{\tabcolsep}{4.5pt}
\renewcommand{\arraystretch}{1.1}
\begin{tabular}{@{}ll rrrrr@{}}
\toprule
\textbf{Model} & \textbf{Paradigm} & \textbf{Prompt} & \textbf{Compl.} & \textbf{Total} & \textbf{Steps} & \textbf{Success Rate} \\
\midrule
\multirow{3}{*}{DeepSeek-V3.2}
 & \cellcolor{highlightcolor}\textbf{CaveAgent} & \cellcolor{highlightcolor}444{,}679 & \cellcolor{highlightcolor}59{,}440 & \cellcolor{highlightcolor}504{,}119 & \cellcolor{highlightcolor}145 & \cellcolor{highlightcolor}\textbf{100\%} \\
 & JSON FC                                       & 660{,}588 & 43{,}600 & 704{,}188 & 236 & 94.6\% \\
 & Bash (filesystem)                             & 1{,}646{,}476 & 80{,}589 & 1{,}727{,}065 & 322 & 98\% \\
\midrule
\multirow{3}{*}{Qwen3-Coder (30B)}
 & \cellcolor{highlightcolor}\textbf{CaveAgent} & \cellcolor{highlightcolor}377{,}599 & \cellcolor{highlightcolor}39{,}111 & \cellcolor{highlightcolor}416{,}710 & \cellcolor{highlightcolor}129 & \cellcolor{highlightcolor}\textbf{83\%} \\
 & JSON FC                                       & 415{,}204 & 15{,}283 & 430{,}487 & 132 & 48\% \\
 & Bash (filesystem)                             & 660{,}378 & 28{,}760 & 689{,}138 & 183 & 62\% \\
\midrule
\multirow{3}{*}{Gemini 3 Pro}
 & \cellcolor{highlightcolor}\textbf{CaveAgent} & \cellcolor{highlightcolor}251{,}774 & \cellcolor{highlightcolor}36{,}327 & \cellcolor{highlightcolor}288{,}101 & \cellcolor{highlightcolor}99 & \cellcolor{highlightcolor}\textbf{96\%} \\
 & JSON FC                                       & 327{,}574 & 36{,}659 & 364{,}233 & 120 & 88\% \\
 & Bash (filesystem)                             & 404{,}860 & 34{,}419 & 439{,}279 & 127 & \textbf{96\%} \\
\bottomrule
\end{tabular}
\end{table}

\paragraph{Results on DeepSeek-V3.2.}
CaveAgent achieves 28.4\% lower total token consumption (504K vs.\ 704K JSON FC) while improving success rate from 94.6\% to 100\%; the gain stems from resolving multiple dependencies in single code executions, reducing steps from 236 to 145 and prompt tokens by 32.7\%. CaveAgent consumes 36.3\% more completion tokens (code is more verbose than JSON), but prompt tokens dominate overall consumption and accumulate across turns. The bash filesystem baseline confirms that the runtime is not over-engineering: filesystem persistence reaches a comparable 98\% success rate but at \textbf{3.7$\times$ the prompt cost} (1.65M vs.\ 445K) and 2.2$\times$ the LLM-call count (322 vs.\ 145), with the surplus traceable not to filesystem I/O itself but to the per-call system-prompt overhead required to teach the dump/load protocol on every API call (decomposition below).

\paragraph{Cross-model directional claim.}
The qualitative finding --- CaveAgent reduces prompt tokens at parity-or-better success rate --- reproduces on Qwen3-Coder (30B) and Gemini~3~Pro. Aggregate prompt savings of CaveAgent over JSON-based function calling are 32.7\% on DeepSeek-V3.2, 23.1\% on Gemini~3~Pro, and 9.1\% on Qwen3-Coder. The shrinking aggregate gap on Qwen is an artifact of \textit{premature exit}: JSON FC succeeds on only 48\% of Qwen scenarios, terminating before its prompts accumulate, which deflates its denominator. Normalising by the count of \textit{successfully completed} turns reverses the impression: CaveAgent's per-success-turn prompt savings are $+37\%$ (DeepSeek), $+30\%$ (Gemini), and $+48\%$ (Qwen) --- the largest savings appear on the weakest model.

\paragraph{Cross-model robustness.}
Across the three models, CaveAgent's success rate spans 17 percentage points (100\%, 96\%, 83\%), against 46.6\,pp for JSON-based function calling (94.6\%, 88\%, 48\%) and 36\,pp for the bash filesystem baseline (98\%, 96\%, 62\%). Within this directional study (single seed, three models), CaveAgent is therefore the least sensitive of the three paradigms to model substitution; larger-$n$ replication beyond the API-cost budget of this revision would be needed to make this a statistical claim. The runtime-mediated mechanism --- variables persist as native objects rather than being re-serialised through the prompt or the filesystem at each turn --- degrades gracefully under weaker models, while both message-passing and filesystem-protocol following amplify the weak-model penalty.

\paragraph{Why bash costs more: a per-call prompt-tax decomposition.}
On the Finance domain (where the gap is most visible on Gemini~3~Pro), bash uses 184K prompt tokens versus CaveAgent's 90K. The 94K excess decomposes into two terms: a system-prompt term (calls~$\times$~instruction size) and a conversation-history term. The system-prompt term contributes 137K to bash and 23K to Cave --- bash carries roughly 2{,}400 tokens of always-resident dump/load instructions per call, against Cave's 580. The conversation-history term is in fact \textit{lower} for bash (47K) than for CaveAgent (67K), since filesystem persistence keeps tool results on disk rather than in the dialogue. The mechanism cost of CaveAgent's runtime is therefore one-time runtime infrastructure that amortises away the per-call instruction overhead a filesystem agent must repeat on every API call.\footnote{The two added models follow the bash system prompt very differently: Qwen3-Coder writes files in 41\% of bash commands (literally following the dump/load recommendation), while Gemini~3~Pro writes files in only 8\% and instead batches several tool invocations into single Python calls in 67\% of bash commands. CaveAgent avoids this strategy choice by construction --- variables persist across turns at zero per-call cost --- so the same architectural mechanism produces uniform behaviour across model families.}

\subsection{[\textbf{Q4}] Case Study: Data-intensive Scenario}

\begin{table}[t]
    \centering
    \caption{Performance comparison across three task categories, framed as a partial ablation of CaveAgent's two principal architectural components: \textit{CodeAct Style} is CaveAgent with the variable injection/retrieval API removed (functionally equivalent to CodeAct's interface), and \textit{JSON-based FC} additionally removes code execution. \textbf{CaveAgent} (highlighted in green) achieves superior performance. Improvements relative to the best baseline are marked in parentheses.}
    \resizebox{\textwidth}{!}{%
        \begin{tabular}{llcccc}
            \toprule
            \textbf{Task Category} & \textbf{Method} & \textbf{Success Rate} {\color{blue}$\uparrow$} & \textbf{Prompt Tokens} {\color{red}$\downarrow$} & \textbf{Compl. Tokens} {\color{red}$\downarrow$} & \textbf{Total Tokens} {\color{red}$\downarrow$} \\
            \midrule

            \multirow{3}{*}{Data Query}
             & \cellcolor{highlightcolor}\textbf{CaveAgent} & \cellcolor{highlightcolor}\textbf{100.0\%} \imp{+20\%} & \cellcolor{highlightcolor}118,901 & \cellcolor{highlightcolor}4,584 & \cellcolor{highlightcolor}123,485 \imp{-51\%} \\
             & CodeAct Style & 80.0\% & 232,990 & 17,219 & 250,209 \\
             & JSON-based FC & 80.0\% & 278,239 & 16,413 & 294,652 \\
            \midrule

            \multirow{3}{*}{Data Analysis}
             & \cellcolor{highlightcolor}\textbf{CaveAgent} & \cellcolor{highlightcolor}\textbf{100.0\%} \imp{Tie} & \cellcolor{highlightcolor}110,550 & \cellcolor{highlightcolor}5,832 & \cellcolor{highlightcolor}116,382 \imp{-2\%} \\
             & CodeAct Style & 100.0\% & 112,990 & 6,232 & 119,222 \\
             & JSON-based FC & 10.0\% & 1,328,779 & 8,024 & 1,336,803 \\
            \midrule

            \multirow{3}{*}{Visualization}
             & \cellcolor{highlightcolor}\textbf{CaveAgent} & \cellcolor{highlightcolor}\textbf{90.0\%} \imp{+50\%} & \cellcolor{highlightcolor}374,855 & \cellcolor{highlightcolor}30,250 & \cellcolor{highlightcolor}405,105 \imp{-39\%} \\
             & CodeAct Style & 40.0\% & 957,447 & 43,144 & 1,000,591 \\
             & JSON-based FC & 30.0\% & 644,778 & 17,899 & 662,677 \\
            \bottomrule
        \end{tabular}%
    }
    \label{data-intensive}
\end{table}

We evaluate three architectures on a data-intensive benchmark comprising 30 tasks across data query, analysis, and visualization using stock market data (Apple and Google, 2020--2025). The setup also serves as a \textbf{partial ablation of CaveAgent's two principal architectural components}: \textit{CodeAct Style} disables CaveAgent's variable injection/retrieval API while retaining the persistent code-execution kernel (its success/failure isolates the contribution of inject/retrieve); \textit{JSON-based Function Calling} additionally removes code execution (isolating the joint contribution of code execution and inject/retrieve relative to the JSON-payload baseline). Results are shown in Table~\ref{data-intensive}.

\paragraph{Results.} On \textit{Data Query}, CaveAgent achieved 100\% accuracy (123K tokens) by storing results in runtime variables, while both baselines failed at 80\% due to context overflow from serializing large datasets. On \textit{Data Analysis}, CaveAgent and CodeAct both achieved 100\% with comparable tokens ($\sim$116--119K), but Function Calling managed only 10\% (1.3M tokens) without code execution. On \textit{Visualization}, CaveAgent achieved 90\% (405K tokens) by retrieving chart data from runtime variables; CodeAct reached 40\% (1M tokens) and Function Calling 30\% (662K tokens). These results demonstrate that decoupling intermediate state from prompt context avoids the token accumulation causing context overflow in conventional architectures, with advantages growing with task complexity and data volume.

\paragraph{Ablation reading.} Read as an ablation, the table isolates the contribution of CaveAgent's two principal components. Removing the inject/retrieve API (\textit{CodeAct Style}) costs \textbf{20 percentage points} on \textit{Data Query} ($100\% \to 80\%$ success) and \textbf{50 points} on \textit{Visualization} ($90\% \to 40\%$): the API is what lets the agent reference large datasets and chart data by handle rather than serializing them into context. Further removing code execution (\textit{JSON-based FC}) is catastrophic on \textit{Data Analysis} (10\% success, 1.3M tokens), confirming that code execution is essential when intermediate computation is data-intensive. This is a partial ablation over two components; ablations of the remaining architectural elements (persistent state vs.\ ephemeral kernel, the Agent-Skills--compatible extension layer, and the multi-agent shared-runtime mode) require separate experimental setups and are left for future work.

\paragraph{Tracing the abstract's ``up to 51\%'' claim.}
Throughout this section, we follow Table~\ref{data-intensive}'s convention and compute per-task token reductions relative to the \textit{best (i.e., most token-efficient) baseline} for that task: CodeAct Style on \textit{Data Query} (250K) and \textit{Data Analysis} (119K), and JSON-based Function Calling on \textit{Visualization} (662K). The largest reduction across the three tasks is on \textit{Data Query}, where CaveAgent's 123K total tokens are 51\% below CodeAct Style's 250K --- this is the ``up to 51\%'' figure quoted in the abstract, and corresponds directly to the \texttt{(\textendash 51\%)} annotation in Table~\ref{data-intensive}. We deliberately do \textit{not} cite CaveAgent's 91\% reduction over JSON-based Function Calling on \textit{Data Analysis} (116K vs.\ 1.3M tokens), even though it is numerically larger: that baseline's 1.3M-token consumption coupled with its 10\% task-success rate jointly indicates a \textit{context-overflow failure mode} --- repeated re-serialization of large DataFrames into the prompt window --- rather than a working baseline whose token cost reflects an actual solution attempt. Reporting a reduction against a failure-mode trajectory would inflate the headline figure, so we exclude this comparison from ``up to'' claims while still reporting the underlying numbers transparently in Table~\ref{data-intensive}. (By contrast, the lower success rates on \textit{Visualization} reflect task difficulty rather than context overflow --- both baselines consume comparable token budgets --- so we retain those comparisons.)


\section{Application Scenarios and Deployment Considerations}
\label{sec:applications}

While Section~\ref{exp} evaluates CaveAgent on standard benchmarks, a reader may reasonably ask how the framework translates into deployed AI/ML systems. This section consolidates four representative scenarios drawn from the qualitative case studies in the appendix --- each generalized from a single demonstration to a class of applications --- and then discusses deployment considerations not surfaced by benchmark evaluation alone (memory budgeting, tool-wrapping cost, auditability, and selection guidance).

\subsection{Application Scenarios}
\label{sec:applications:scenarios}

\paragraph{Stateful device control.} The \textit{Smart Home} case study (Appendix~\ref{features_section}, Figure~\ref{fig:smart_home_demo}) shows how the dual-stream split substitutes for hand-coded state machines in rule-based home-automation AI/ML systems. Variables such as \texttt{Thermostat} and \texttt{Door} are initialized once and persist across turns; the agent generates Python conditionals (e.g., \texttt{if not door\_lock.is\_locked:}) rather than blind API calls. The same pattern fits any IoT or supervisory-control AI/ML system where (i) device state must remain consistent across user interactions and (ii) decision logic is more naturally expressed as code than as a fixed rule table --- an alternative deployment style to engines such as openHAB or Home Assistant rule chains.

\paragraph{Scientific decision support over non-serializable inputs.} The \textit{Geospatial Analysis} case study (Appendix~\ref{app:geospatial}) illustrates a pattern we expect to recur in scientific AI/ML systems: the input is a complex non-textual object (here, GeoJSON polygons with high-precision floating-point coordinates) that loses precision or fails entirely when serialized to a JSON payload. CaveAgent injects the polygon as a first-class Python variable and resolves the spatial query in a single turn against domain libraries such as \texttt{osmnx}; the same task under JSON-based function calling requires at least five sequential turns and risks coordinate truncation. The same property generalizes to medical imaging (NIfTI volumes), computational geometry / CAD (mesh objects), and bioinformatics (BioPython records) --- application classes where ML practitioners face the same serialization wall.

\paragraph{Hierarchical pipeline orchestration.} The \textit{AutoML Training Loop} (Appendix~\ref{app:automl}, Figure~\ref{fig:automl}) exhibits a different reuse pattern: an orchestrator agent injects raw data into a feature-engineering sub-agent's runtime via \texttt{inject()}, retrieves the transformed DataFrame, and forwards it to a trainer sub-agent --- all without serialization between stages. This is the structure of typical MLOps pipelines, in which adapter code between stages is often the dominant integration cost; CaveAgent's typed bidirectional flow replaces such adapters with native Python object handoff and supports automated convergence checks via inspection of the trainer's runtime metrics.

\paragraph{Multi-agent shared-world simulation.} The \textit{Town Simulation} (Appendix~\ref{app_multiagent}, Figure~\ref{Town}) demonstrates peer-to-peer coordination through a shared runtime: when the meta-agent modifies a global \texttt{weather} entity, all resident agents observe the change through direct attribute access rather than through inter-agent messaging. This is the structural pattern behind digital-twin and agent-based simulation AI/ML systems (city modeling, supply-chain simulation, epidemiological models) --- domains in which message-passing implementations are prone to message-ordering ambiguity on shared world state. Under CaveAgent's single-threaded, turn-based runtime, state updates are linearized through the kernel namespace, which avoids that class of bug at the cost of forgoing concurrent execution between agents.

\subsection{Deployment Considerations}
\label{sec:applications:deployment}

\paragraph{Memory budget and cold start.} The persistent IPython kernel carries a baseline memory footprint plus payload proportional to the size of injected and retained objects, and the first turn of each session pays a kernel-boot cost not incurred by stateless JSON function calling. We hypothesize two mitigations as future systems work rather than measured deployment guidance: a kernel pool that amortizes cold-start across requests for high-throughput deployments, and explicit \texttt{del} of intermediate variables or session checkpointing for long sessions to limit payload growth. Cold-start, throughput, and production-scale memory profiling were not in the scope of this paper; both mitigations are testable in any specific deployment.

\paragraph{Tool wrapping at the Python boundary.} As acknowledged in our Limitations, CaveAgent's design couples execution to a Python interpreter; tools exposed only through non-Python interfaces require wrapper creation. In practice the wrapper is small: a tool reachable via REST or gRPC is typically wrapped in a few lines of \texttt{requests} or grpc client code, and an in-process Java/Go service can be exposed through a thin RPC shim. ML practitioners with substantial existing capability in other-language services therefore face a per-tool integration cost rather than a wholesale rewrite.

\paragraph{Auditability and runtime-state inspection.} CaveAgent's static-analysis security checks (\textit{ImportRule}, \textit{FunctionRule}, \textit{AttributeRule}) filter dangerous code at execution time; complementarily, the runtime namespace is post-hoc inspectable, so any variable created or modified during a session can be examined for compliance review or fault diagnosis. This is a structural property of the architecture rather than a fully tooled deliverable: comprehensive audit-grade tooling (chain-of-custody logs, immutable run histories) remains future work, but the underlying inspectability is what regulated-deployment use cases (healthcare decision support, financial advisory) typically require.

\paragraph{Selection guidance.} CaveAgent is the appropriate choice when an application combines several of: multi-turn state spanning many turns, non-serializable or precision-sensitive inputs, multi-agent state handoff, or audit trails over intermediate state. Conversely, JSON-based function calling remains preferable for short single-turn queries where memory footprint and cold-start latency dominate, and CodeAct's text-bound interface is sufficient when persistence across sessions is not desired (for example, fully ephemeral execution environments). We recommend treating CaveAgent as one design point among these alternatives rather than a uniform replacement.

\subsection{Implications for Deployed ML Systems}
\label{sec:applications:benefits}

\paragraph{Programmatic verifiability and audit trails.} Because runtime state is a deterministically inspectable Python namespace, agent behavior can be evaluated without relying on subjective human annotation --- a property already noted as a foundation for reinforcement learning with verifiable rewards. Reframed for deployment, the same property supports compliance review in regulated domains: every variable created or modified during a session can be queried after the fact, enabling fine-grained audit trails over intermediate variable state. Call-level logging under JSON-based function calling provides only the surface trace of tool invocations and their textual outputs and does not natively expose the intermediate runtime state from which those outputs were derived.

\paragraph{Skill portability via the Agent Skills standard.} CaveAgent's \texttt{injection.py} extension to the Agent Skills standard (Section~\ref{sec:skills}) allows domain expertise --- medical-coding rules, financial-product validators, geospatial analysis recipes --- to be packaged as portable, versionable artifacts. This is the modern analogue of the rule packs traditionally distributed for expert-system shells such as CLIPS or JESS, with the difference that the skill itself contributes executable functions and typed objects to the runtime, not only natural-language instructions.

\paragraph{Lossless object handoff to downstream pipelines.} Agents return native Python objects rather than text approximations, allowing downstream consumers (BI dashboards, training harnesses, automated test rigs) to bind to results directly rather than re-parsing serialized output. This avoids a serialize / deserialize round-trip whose cost, on data-intensive tasks, accounted for substantial overhead under JSON-based function calling (Section~\ref{exp}).


\section{Conclusion}
\label{conc}
We present \textbf{CaveAgent}, a framework for LLM tool use based on persistent, object-oriented stateful runtime management as an alternative to stateless JSON function calling. CaveAgent enables agents to maintain high-fidelity memory of complex objects and execute sophisticated logic via Python code. By extending the Agent Skills open standard with runtime injection, CaveAgent demonstrates that the persistent runtime paradigm enables a new mode of tool distribution, skills that deliver executable artifacts rather than solely textual instructions, unifying tool registration and skill management into a single portable mechanism. Experiments on Tau$^2$-bench show that this approach consistently outperforms SOTA baselines in multi-turn success rates (11 out of 12 settings) and token efficiency. On BFCL, the three open-source models we evaluate (DeepSeek-V3.2, Qwen3-Coder 30B, and Kimi-K2) all reach 94.0--94.7\% under CaveAgent, comparable to the closed-source Claude Sonnet 4.5 (94.4\%) and Gemini 3 Pro (94.3\%) and exceeding GPT-5.1 (89.6\%) under their native function-calling protocols; the 30B Qwen3-Coder matching Claude Sonnet 4.5 (both at 94.4\%) further suggests that for code-capable LLMs the function-calling protocol can be as significant a performance bottleneck as model scale. Beyond performance gains, a key contribution is the \textbf{programmatic verifiability} enabled by CaveAgent's architecture: because runtime state is deterministically inspectable, agent behavior can be evaluated automatically without human annotation, establishing a structural foundation for Reinforcement Learning with Verifiable Rewards and runtime-mediated multi-agent coordination. Qualitative case studies are provided in Appendix \ref{features_section}.

\paragraph{Limitations.} \textit{Reliance on a Python runtime.} CaveAgent's design fundamentally couples the agent's execution to a Python interpreter: tools must be Python-callable or wrapped as Python functions, and the persistent state lives in a Python kernel namespace. Tools exposed only through non-Python interfaces (e.g., native binaries, proprietary REST APIs in other languages, microservices written in Java/Go/Rust) require manual wrapper creation, and the runtime state cannot be transparently shared with non-Python downstream systems --- a generalizability bound intrinsic to this design choice rather than an implementation gap. Extending the dual-stream pattern to language-agnostic runtimes (e.g., WebAssembly-based execution that admits multi-language tool implementations) is a natural future direction. \textit{Memory.} The persistent runtime consumes memory proportional to the complexity of stored objects; for extremely long sessions with large data artifacts, memory management becomes a concern. \textit{Stateful evaluation.} While agent behavior is in principle programmatically verifiable, designing comprehensive fine-grained benchmarks for stateful evaluation remains future work. \textit{Multi-agent.} The multi-agent coordination capabilities are demonstrated qualitatively; rigorous quantitative evaluation of runtime-mediated multi-agent systems is left for future investigation. \textit{Q3 multi-model and bash-baseline scope.} The two added model families and the bash filesystem paradigm in Table~\ref{tab:Q3_aggregate} are at $n{=}1$ per cell; we report directional findings rather than statistical significance and recommend $n{\geq}3$ replication for any specific deployment decision. Extending the bash filesystem baseline to the Q4 data-intensive setup is left for future work --- Q4's validators inspect runtime state directly and would need parallel file-fallback paths, and Q3's small structured-dict data is in fact the configuration most favourable to a filesystem agent, so the runtime-versus-filesystem gap on Q4 is expected to be larger, not smaller.


\section{Acknowledgment}
\begin{itemize}[itemsep=0.3cm]
    \item We thank Rui Zhou, a professional UI designer at Metasequoia Tech, for his assistance with the figure design in this paper.
    \item We thank Qiuyang Mang, a Ph.D student in Computer Science at UC Berkeley, for the discussion about the core design of our framework.
\end{itemize}

\bibliography{main}  

\newpage
\appendix
\newpage

\begin{center}
{\Large \textbf{Appendix}}
\end{center}
\vspace{1em}
    
{\hypersetup{linktoc=page}
\startcontents[appendix]
\printcontents[appendix]{l}{1}{\setcounter{tocdepth}{2}}}
    
\clearpage
    

\definecolor{promptbg}{RGB}{245,245,245}
\definecolor{promptframe}{RGB}{200,200,200}

\section{Pseudo Code}
\label{algo_appendix}
Algorithm \ref{alg:cave_agent} shows the general workflow of CaveAgent.
\begin{algorithm}[t]
\caption{CaveAgent Interaction Loop}
\label{alg:cave_agent}
\begin{algorithmic}[1]
\Require Query $q$, Tools $\mathcal{T}$, Max Turns $T_{\max}$
\State $\mathcal{S}_0 \leftarrow$ \textsc{InitKernel}(); \textsc{Inject}($\mathcal{S}_0, \mathcal{T}$) \Comment{Init Runtime Stream}
\State $D \leftarrow$ \textsc{GenSigs}($\mathcal{T}$); $H_0 \leftarrow \{ \text{Sys}(D), \text{User}(q) \}$ \Comment{Init Semantic Stream}

\For{$t = 1$ to $T_{\max}$}
    \State \textbf{Phase 1 (Reasoning):} $R_t \leftarrow \text{LLM}(H_{t-1})$ \Comment{Sample thought \& code}
    
    \If{$R_t$ contains code block $c_t$}
        \State \textbf{Phase 2 (Security):} $V \leftarrow \textsc{ASTCheck}(c_t, \Pi)$ \Comment{Pre-exec validation}
        \If{$V \neq \emptyset$}
            \State $o_t \leftarrow \text{FormatError}(V)$
        \Else
            \State \textbf{Phase 3 (Execution):} $o_t, \mathcal{S}_t \leftarrow \textsc{Run}(\mathcal{S}_{t-1}, c_t)$ \Comment{Stateful update}
            \State \textbf{Phase 4 (Shaping):} $o_t \leftarrow \textsc{Shape}(o_t, L_{\max})$ \Comment{Truncate \& format}
        \EndIf
        \State $H_t \leftarrow H_{t-1} \cup \{ (R_t, o_t) \}$ \Comment{Sync observation to history}
    \Else
        \State \textbf{return} $R_t$ \Comment{Output final answer}
    \EndIf
\EndFor
\State \textbf{return} "Max steps reached"
\end{algorithmic}
\end{algorithm}

\begin{figure}[!htbp]
    \centering
    \includegraphics[width=\linewidth]{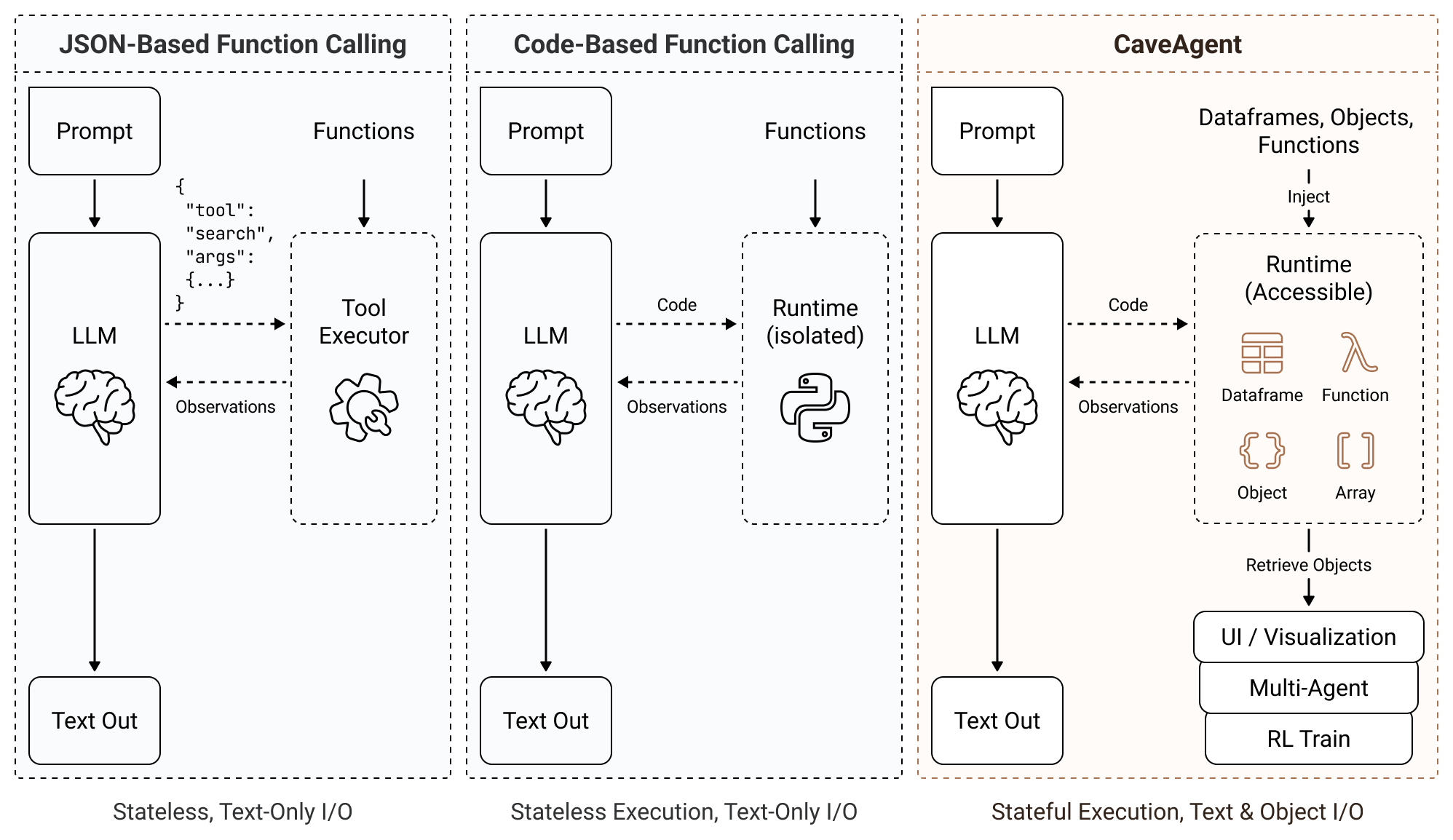}
    \caption{Evolution of Agentic Tool Use}
    \label{evolution}
\end{figure}

\section{What Happens in Semantic Stream}
\label{app:prompts}

The following sections detail the prompt templates used to instruct the Semantic Stream in CaveAgent. The system prompt is dynamically constructed by combining the Agent Identity, Context Information (functions, variables, types), and Instructions.

\subsection{System Prompt Construction}

The full system prompt is composed using the following template structure. The placeholders (e.g., \texttt{\{functions\}}) are populated at runtime with the specific tools and variables available in the current environment.

\begin{tcolorbox}[
    breakable,
    enhanced,
    colback=promptbg,
    colframe=promptframe,
    title=\textbf{System Prompt Template},
    fonttitle=\bfseries\sffamily,
    arc=2mm,
    boxrule=1pt,
    left=10pt, right=10pt, top=10pt, bottom=10pt
]
\texttt{\{agent\_identity\}}

\vspace{1em}
Current time: \texttt{\{current\_time\}}

\vspace{1em}
You have access to:

\vspace{0.5em}
<functions>\\
\texttt{\{functions\}}\\
</functions>

\vspace{0.5em}
<variables>\\
\texttt{\{variables\}}\\
</variables>

\vspace{0.5em}
<types>\\
\texttt{\{types\}}\\
</types>

\vspace{1em}
Instructions:\\
\texttt{\{instructions\}}

\vspace{1em}
\texttt{\{additional\_context\}}
\end{tcolorbox}

\vspace{1em}
Below are the default values for the key components referenced in the template above.

\begin{tcolorbox}[
    colback=promptbg,
    colframe=promptframe,
    title=\textbf{Component: Agent Identity},
    fonttitle=\bfseries\sffamily,
    arc=2mm,
    boxrule=1pt
]
You are a tool-augmented agent specializing in Python programming that enables function-calling through LLM code generation. You have to leverage your coding capabilities to interact with tools through a Python runtime environment, allowing direct access to execution results and runtime state. The user will give you a task and you should solve it by writing Python code in the Python environment provided.
\end{tcolorbox}

\begin{tcolorbox}[
    colback=promptbg,
    colframe=promptframe,
    title=\textbf{Component: Core Instructions},
    fonttitle=\bfseries\sffamily,
    arc=2mm,
    boxrule=1pt
]
1. Carefully read and analyze the user's input.\\

2. If the task requires Python code:
   - Generate appropriate Python code to address the user's request.
   - Your code will then be executed in a Python environment, and the execution result will be returned to you as input for the next step.
   - During each intermediate step, you can use 'print()' to save whatever important information you will then need in the following steps.
   - These print outputs will then be given to you as input for the next step.
   - Review the result and generate additional code as needed until the task is completed.\\
   
3. CRITICAL EXECUTION CONTEXT: You are operating in a persistent Jupyter-like environment where:
  - Each code block you write is executed in a new cell within the SAME continuous session
  - ALL variables, functions, and imports persist across cells automatically
  - You can directly reference any variable created in previous cells without using locals(), globals(), or any special access methods.\\
  
4. If the task doesn't require Python code, provide a direct answer based on your knowledge.\\

5. Always provide your final answer in plain text, not as a code block.\\

6. You must not perform any calculations or operations yourself, even for simple tasks like sorting or addition. \\

7. Write your code in a \texttt{\{python\_block\_identifier\}} code block. In each step, write all your code in only one block.\\

8. Never predict, simulate, or fabricate code execution results.\\

9. To solve the task, you must plan forward to proceed in a series of steps, in a cycle of Thought and Code sequences.\\

\end{tcolorbox}

\subsection{Context Injection Format}
\label{app:context_injection}

Examples of how context is formatted for the LLM.

\begin{tcolorbox}[
    breakable,
    enhanced,
    colback=promptbg,
    colframe=promptframe,
    title=\textbf{Example: Function Injection},
    fonttitle=\bfseries\sffamily,
    arc=2mm,
    boxrule=1pt
]
<functions>\\
- function: buy\_stock(symbol: str, quantity: int) -> Transaction\\
  description: Executes a stock purchase for the current portfolio.\\
  doc:\\
\phantom{xx}Args:\\
\phantom{xxxx}symbol: The ticker symbol of the stock (e.g., 'AAPL').\\
\phantom{xxxx}quantity: The number of shares to purchase.\\
\phantom{xx}Returns:\\
\phantom{xxxx}A Transaction object recording the details of the purchase.\\
</functions>
\end{tcolorbox}

\begin{tcolorbox}[
    colback=promptbg,
    colframe=promptframe,
    title=\textbf{Example: Variable Injection},
    fonttitle=\bfseries\sffamily,
    arc=2mm,
    boxrule=1pt
]
<variables>\\
- name: portfolio\\
  type: Portfolio\\
  description: The user's current investment portfolio object.\\
\\
- name: market\_data\\
  type: DataFrame\\
  description: A pandas DataFrame containing historical price data.\\
</variables>
\end{tcolorbox}

\begin{tcolorbox}[
    colback=promptbg,
    colframe=promptframe,
    title=\textbf{Example: Type Schema Injection},
    fonttitle=\bfseries\sffamily,
    arc=2mm,
    boxrule=1pt
]
<types>\\
Portfolio:\\
\phantom{xx}doc: Manages a collection of stock holdings and cash balance.\\
\phantom{xx}methods:\\
\phantom{xxxx}- get\_total\_value() -> float\\
\phantom{xxxx}- get\_holdings() -> Dict[str, int]\\
\phantom{xxxx}- add\_cash(amount: float) -> None\\
\\
Transaction:\\
\phantom{xx}doc: An immutable record of a stock transaction.\\
\phantom{xx}fields:\\
\phantom{xxxx}- id: str\\
\phantom{xxxx}- symbol: str\\
\phantom{xxxx}- quantity: int\\
\phantom{xxxx}- price\_at\_execution: float\\
\phantom{xxxx}- timestamp: datetime\\
</types>
\end{tcolorbox}

\subsection{Runtime Feedback Prompts}
\label{app:feedback}

The agent operates in a closed feedback loop. After each code execution step, the runtime environment captures the output (stdout or errors) and constructs a new user message to guide the agent's next action.

\subsubsection{Standard Execution Output}
This prompt is used when code executes successfully. It provides the standard output and explicitly reminds the agent that the variable state has been preserved.

\begin{tcolorbox}[
    breakable,
    enhanced,
    colback=promptbg,
    colframe=promptframe,
    title=\textbf{Execution Output Template},
    fonttitle=\bfseries\sffamily,
    arc=2mm,
    boxrule=1pt
]
<execution\_output>\\
\texttt{\{execution\_output\}}\\
</execution\_output>
\\
\\
IMPORTANT CONTEXT REMINDER:
- Based on this output, should we continue with more operations? \\
- If the output includes an error, please review the error carefully and modify your code to fix the error if needed.\\
- If yes, provide the next code block. If no, provide the final answer (not as a code block).\\
- You are in the SAME Jupyter-like session. All variables from your previous code blocks are still available and can be accessed directly by name.\\
- You DO NOT need to use locals(), globals(), or any special methods to access them.\\
- Think of this exactly like working in Jupyter: when you create a variable in cell 1, you can simply use it by name in cell 2, 3, 4, etc.
\end{tcolorbox}

\subsubsection{Error Handling \& Constraints}
The system includes specific templates for handling edge cases, such as context window limits and security violations.

\textbf{Output Length Exceeded:}
Used when the code generates excessive output (e.g., printing a massive DataFrame), prompting the agent to summarize instead.

\begin{tcolorbox}[
    colback=promptbg,
    colframe=promptframe,
    title=\textbf{Output Truncation Template},
    fonttitle=\bfseries\sffamily,
    arc=2mm,
    boxrule=1pt
]
The code execution generated {output\_length} characters of output, which exceeds the maximum limit of {max\_length} characters.
Please modify your code to:\\
1. Avoid printing large datasets or lengthy content\\
2. Use summary statistics instead of full data (e.g., print shape, head(), describe() for dataframes)\\
3. Print only essential information needed for the task
"""
\end{tcolorbox}

\textbf{Security Violation:}
Used when the static analysis security checker blocks unsafe code (e.g., \texttt{os.system}).

\begin{tcolorbox}[
    colback=promptbg,
    colframe=promptframe,
    title=\textbf{Security Error Template},
    fonttitle=\bfseries\sffamily,
    arc=2mm,
    boxrule=1pt
]
<security\_error>\\
\texttt{\{error\}}\\
</security\_error>\\
Code blocked for security reasons. Please modify your code to avoid this violation.
\end{tcolorbox}

\section{What Happened in Runtime Stream}
\label{sec:runtime_stream}

While the Semantic Stream governs reasoning and planning, the \textbf{Runtime Stream} serves as the execution engine and persistent memory. This stream operates as a dedicated Python kernel where data manipulation, tool invocation, and state transitions occur. The two streams follow a strict chronological topology, synchronized through interleaved exchange of code instructions and execution feedback.

\subsection{Environment Initialization via Injection}
The runtime lifecycle begins with \textbf{Context Injection}. Before the reasoning cycle starts, the user (or the system orchestration layer) initializes the runtime environment by injecting native Python objects directly into the global namespace.
\begin{itemize}
    \item \textbf{Function Injection:} Tool definitions are loaded as executable Python callables. Unlike RESTful API wrappers, these are native functions that can be inspected and invoked directly.
    \item \textbf{Variable Injection:} Domain-specific data, such as DataFrames, graph structures, or class instances, are instantiated within the runtime stream's memory.
\end{itemize}
This initialization phase populates the \texttt{<functions>} and \texttt{<variables>} blocks described in Section \ref{app:prompts}.

\subsection{The Interleaved Execution Paradigm}
Once initialized, the workflow proceeds as a synchronized dialogue between the Semantic Stream (Reasoning) and the Runtime Stream (Execution). We conceptualize this as a dual-column timeline where actions are interleaved strictly in chronological order:

\begin{enumerate}
    \item \textbf{Semantic Turn (Left Cell):}
    The LLM analyzes the current task and available context. It generates a \textit{Thought} followed by a discrete \textit{Code Block} (the instruction). This represents the input to the runtime.
    
    \item \textbf{Runtime Turn (Right Cell):}
    The system extracts the code block and executes it within the persistent Python kernel. This execution constitutes the state transition $S_t \rightarrow S_{t+1}$. Crucially, this is not a stateless function call; it is a stateful operation where:
    \begin{itemize}
        \item New variables defined in this cell are persisted in memory.
        \item Existing objects (e.g., a list or a database connection) are mutated in place.
        \item Side effects (e.g., saving a file) are realized immediately.
    \end{itemize}
    
    \item \textbf{Feedback Loop:}
    Upon completion of the Runtime Turn, the standard output (\texttt{stdout}), standard error (\texttt{stderr}), or the return value of the last expression is captured. This raw execution result is wrapped in the \texttt{<execution\_output>} tags and injected back into the Semantic Stream, triggering the next Semantic Turn.
\end{enumerate}

This mechanism ensures that the agent's reasoning is always grounded in the current, actual state of the runtime environment.

\subsection{Illustrative Case Study}
\label{sec:case_study}

To intuitively demonstrate the temporal synchronization and state dependency between the two streams, we present a concrete walkthrough in Figure \ref{fig:interaction_trace}. This example illustrates a toy data analysis task where the agent must filter a dataset and perform calculations on the result.

The workflow proceeds in a ``zig-zag'' pattern, alternating between reasoning (Left) and execution (Right):

\begin{enumerate}
    \item \textbf{Initialization ($T_0$):} The user injects a pandas DataFrame named \texttt{df}. Note that the Semantic Stream only receives a lightweight pointer (variable name and documentation) instead of the whole data, while the Runtime Stream holds the actual heavy data object in memory.
    \item \textbf{Step 1 ($T_1 \rightarrow T_2$):} The agent generates code to filter the data. Crucially, the Runtime Stream does not return the full filtered dataset as text. Instead, it creates a new variable \texttt{high\_vol} in the local scope and returns only a status update. This exemplifies our \textbf{Stateful Management}: the ``result'' of the tool use is a state change in memory, not a text string.
    \item \textbf{Step 2 ($T_3 \rightarrow T_4$):} The agent references the \textit{previously created} variable \texttt{high\_vol} to compute a statistic. This demonstrates \textbf{Context Compression}: the agent manipulates the data via variable references without ever consuming context tokens to ``read'' the full dataset.
\end{enumerate}

\begin{figure*}[!htbp]
    \centering
    \footnotesize
    \renewcommand{\arraystretch}{1.3}
    \definecolor{sem_bg}{RGB}{245, 247, 250} 
    \definecolor{run_bg}{RGB}{240, 255, 240} 
    \definecolor{code_color}{RGB}{0, 50, 150}
    
    \begin{tcolorbox}[
        colback=white,
        colframe=black!60,
        boxrule=0.8pt,
        title=\textbf{Timeline of Interleaved Execution (Semantic vs. Runtime)},
        fonttitle=\bfseries
    ]
    \begin{tabularx}{\textwidth}{@{}p{0.05\textwidth}XX@{}}
        \midrule
        \textbf{Time} & \cellcolor{gray!20}\textbf{Semantic Stream (Reasoning \& Code)} & \cellcolor{gray!20}\textbf{Runtime Stream (State \& Execution)} \\
        \midrule
        
        $T_0$ & 
        \textit{(Waiting for Input)} & 
        \textbf{[Initialization]} \newline
        $\bullet$ \texttt{df = pd.read\_csv("market.csv")} loaded into memory. \newline
        $\bullet$ \texttt{<variables>} metadata sent to Left Stream. \\
        \midrule
        
        $T_1$ & 
        \cellcolor{sem_bg}
        \textbf{User:} ``Find stocks with volume > 1M.'' \newline
        \textbf{Agent Thought:} I need to filter the dataframe \texttt{df}. I will save the result to a new variable. \newline
        \textbf{Code:} \newline
        \texttt{\textcolor{code_color}{high\_vol = df[df['volume'] > 1e6]}} \newline
        \texttt{\textcolor{code_color}{print(f"Filtered. Rows: \{len(high\_vol)\}")}} & 
        \textit{(Waiting for Instruction)} \\
        \midrule
        
        $T_2$ & 
        \textit{(Waiting for Feedback)} & 
        \cellcolor{run_bg}
        \textbf{[Execution]} \newline
        1. Evaluates filter logic on \texttt{df}. \newline
        2. \textbf{New State:} Variable \texttt{high\_vol} created in RAM. \newline
        3. \textbf{Output:} \texttt{``Filtered. Rows: 42''} sent to Left Stream. \\
        \midrule
        
        $T_3$ & 
        \cellcolor{sem_bg}
        \textbf{Input:} \texttt{``Filtered. Rows: 42''} \newline
        \textbf{Agent Thought:} Now I need to calculate the average price of these high-volume stocks. \newline
        \textbf{Code:} \newline
        \texttt{\textcolor{code_color}{avg\_price = high\_vol['price'].mean()}} \newline
        \texttt{\textcolor{code_color}{print(round(avg\_price, 2))}} & 
        \textit{(Waiting for Instruction)} \\
        \midrule
        
        $T_4$ & 
        \textit{(Waiting for Feedback)} & 
        \cellcolor{run_bg}
        \textbf{[Execution]} \newline
        1. Accesses persistent object \texttt{high\_vol}. \newline
        2. Computes mean. \newline
        3. \textbf{Output:} \texttt{154.20} sent to Left Stream. \\
        \midrule
    \end{tabularx}
    \end{tcolorbox}
    \caption{\label{fig:interaction_trace} A visualization of the CaveAgent workflow. The process alternates between the Semantic Stream (generating instructions) and the Runtime Stream (executing and maintaining state). Note how the variable \texttt{high\_vol} is maintained in the Runtime Stream ($T_2$) and accessed in the subsequent step ($T_4$) without re-loading or serialization, illustrating the efficiency of Stateful Runtime Management.}
\end{figure*}

Viewing the runtime stream as a Jupyter notebook with multiple cells, where each cell corresponds to the execution at each time step, helps illustrate how states remain persistent across steps.

\clearpage
\section{Test Cases in Stateful Management Benchmark}
\label{cases}

In this section, we provide the examples of our test cases in Stateful Management Benchmark.
\subsection{Type Proficiency Cases}
The \textbf{Type Proficiency} category evaluates the agent's ability to perform precise, state-aware manipulation of Python runtime elements. This section tests the agent's working memory across three structural tiers: \textit{Simple Types} (primitive types such as lists, dictionaries, and strings), \textit{Object Types} (custom classes), and \textit{Scientific Types} (high-dimensional complex data). Proficiency in these domains is a prerequisite for complex reasoning tasks.
\subsubsection{Simple Types}
Figure \ref{fig:step_by_step_simple} shows the examples of our test cases of Simple types.

\begin{figure*}[!htbp]
    \centering
    \footnotesize
    \renewcommand{\arraystretch}{1.3}
    
    \definecolor{tbl_header}{RGB}{240, 242, 245}
    \definecolor{turn_label}{RGB}{100, 100, 100}
    \definecolor{val_func}{RGB}{180, 0, 0}      
    \definecolor{val_desc}{RGB}{0, 100, 60}     

    \providecommand{\step}[1]{\textcolor{turn_label}{\textbf{#1}}}
    \providecommand{\validator}[1]{\texttt{\textcolor{val_func}{#1}}}

    \begin{tcolorbox}[
        colback=white,
        colframe=black!70,
        boxrule=0.8pt,
        arc=2pt,
        title=\textbf{Simple Types},
        fonttitle=\bfseries\small
    ]
    
    \begin{tabularx}{\textwidth}{@{}p{0.08\textwidth}XX@{}}
        \midrule
        \rowcolor{tbl_header}
        \textbf{Turn} & \textbf{User Query (Input)} & \textbf{Immediate Validation (State Assertion)} \\
        \midrule

        \multicolumn{3}{@{}l@{}}{\cellcolor{tbl_header!50}\textbf{\textit{Case: \texttt{string\_split\_join}}}} \\
        \midrule
        \step{T1} & 
        ``Set text to 'a,b,c', split by comma, and rejoin with ' | ' as separator...'' & 
        \validator{validate\_str\_split} \newline
        $\bullet$ Assert \texttt{text} == "a | b | c". \\
        \midrule
        \step{T2} & 
        ``Sort the parts of text alphabetically while keeping the ' | ' separator format.'' & 
        \validator{validate\_str\_sort} \newline
        $\bullet$ Assert \texttt{text} remains "a | b | c". \newline
        $\bullet$ Checks persistence of structure. \\
        \midrule
        \step{T3} & 
        ``Reverse the order of parts in text but keep the ' | ' separator...'' & 
        \validator{validate\_str\_reverse} \newline
        $\bullet$ Assert \texttt{text} == "c | b | a". \\
        \midrule

        \multicolumn{3}{@{}l@{}}{\cellcolor{tbl_header!50}\textbf{\textit{Case: \texttt{dict\_nested}}}} \\
        \midrule
        \step{T1} & 
        ``Change the math score to 90 in data['scores']['math'].'' & 
        \validator{validate\_dict\_nested\_update} \newline
        $\bullet$ Assert \texttt{data['scores']['math']} == 90. \\
        \midrule
        \step{T2} & 
        ``The student just took a science test. Add a science score of 88 to data['scores'].'' & 
        \validator{validate\_dict\_nested\_add} \newline
        $\bullet$ Assert key \texttt{'science'} exists with value 88. \\
        \midrule
        \step{T3} & 
        ``There was a curve on all tests. \textbf{Add 5 points to every score} in the scores dictionary.'' & 
        \validator{validate\_dict\_increment} \newline
        $\bullet$ Assert \texttt{math} == 95 (90+5). \newline
        $\bullet$ Assert \texttt{science} == 93 (88+5). \newline
        $\bullet$ Assert \texttt{english} == 95 (Initial 90+5). \\
        \midrule
        
    \end{tabularx}
    \end{tcolorbox}
    \vspace{-0.5em}
    \caption{\label{fig:step_by_step_simple} \textbf{Illustration of test cases of Simple Types.} The results show the agent's capability to manipulate any object types.}
\end{figure*}

\subsubsection{Object Types}
Figure \ref{fig:step_by_step_objective} shows the examples of our test cases of Object types.
\begin{figure*}[!htbp]
    \centering
    \footnotesize
    \renewcommand{\arraystretch}{1.3}
    
    \definecolor{tbl_header}{RGB}{240, 242, 245}
    \definecolor{turn_label}{RGB}{100, 100, 100}
    \definecolor{val_func}{RGB}{180, 0, 0}      
    \definecolor{val_desc}{RGB}{0, 100, 60}     

    \providecommand{\step}[1]{\textcolor{turn_label}{\textbf{#1}}}
    \providecommand{\validator}[1]{\texttt{\textcolor{val_func}{#1}}}

    \begin{tcolorbox}[
        colback=white,
        colframe=black!70,
        boxrule=0.8pt,
        arc=2pt,
        title=\textbf{Object Types},
        fonttitle=\bfseries\small
    ]
    
    \begin{tabularx}{\textwidth}{@{}p{0.08\textwidth}XX@{}}
        \midrule
        \rowcolor{tbl_header}
        \textbf{Turn} & \textbf{User Query (Input)} & \textbf{Immediate Validation (State Assertion)} \\
        \midrule
        
        \multicolumn{3}{@{}l@{}}{\cellcolor{tbl_header!50}\textbf{\textit{Case: \texttt{stack\_advanced}}}} \\
        \midrule
        \step{T1} & 
        ``Push 'A', 'B', 'C', 'D' in order.'' & 
        \validator{validate\_stack\_multi\_push} \newline
        $\bullet$ Assert \texttt{stack.size() == 4}. \\
        \midrule
        \step{T2} & 
        ``User wants to go back to first page. \textbf{Pop until only 1 item remains}, store count in \texttt{result\_num}.'' & 
        \validator{validate\_stack\_pop\_until} \newline
        $\bullet$ Assert \texttt{stack.size() == 1}. \newline
        $\bullet$ Assert \texttt{result\_num == 3} (Popped D,C,B). \\
        \midrule
        \step{T3} & 
        ``Verify we're at the right page. Peek at stack's top and store in \texttt{result\_str}.'' & 
        \validator{validate\_stack\_peek\_after} \newline
        $\bullet$ Assert \texttt{result\_str == 'A'}. \newline
        $\bullet$ Assert \texttt{stack.size() == 1}. \\
        \midrule
        
        \multicolumn{3}{@{}l@{}}{\cellcolor{tbl_header!50}\textbf{\textit{Case: \texttt{cart\_quantity}}}} \\
        \midrule
        \step{T1} & 
        ``Add 3 Apples at \$10.00 each to cart with quantity.'' & 
        \validator{validate\_cart\_qty\_add} \newline
        $\bullet$ Assert \texttt{len(cart.items) == 1}. \newline
        $\bullet$ Assert \texttt{items[0]['quantity'] == 3}. \\
        \midrule
        \step{T2} & 
        ``Also add 2 Oranges at \$5.00 each...'' & 
        \validator{validate\_cart\_qty\_add2} \newline
        $\bullet$ Assert \texttt{len(cart.items) == 2}. \\
        \midrule
        \step{T3} & 
        ``Calculate total (price * quantity)... store in \texttt{result\_num}.'' & 
        \validator{validate\_cart\_qty\_total} \newline
        $\bullet$ Assert \texttt{result\_num == 40.0}. \newline
        $\bullet$ Logic: $(3 \times 10) + (2 \times 5)$. \\
        \midrule

    \end{tabularx}
    \end{tcolorbox}
    \vspace{-0.5em}
    \caption{\label{fig:step_by_step_objective} \textbf{Illustration of test cases of Object Types.} The results show the agent's capability to manipulate custom class instances (Stack, ShoppingCart, Person) and verifying their internal attributes and method side-effects.}
\end{figure*}

\subsubsection{Scientific Types}
Figure \ref{fig:step_by_step_scientific} shows the examples of our test cases of Scientific types.
\begin{figure*}[!htbp]
    \centering
    \footnotesize
    \renewcommand{\arraystretch}{1.3}
    
    \definecolor{tbl_header}{RGB}{240, 242, 245}
    \definecolor{turn_label}{RGB}{100, 100, 100}
    \definecolor{val_func}{RGB}{180, 0, 0}
    
    \providecommand{\step}[1]{\textcolor{turn_label}{\textbf{#1}}}
    \providecommand{\validator}[1]{\texttt{\textcolor{val_func}{#1}}}

    \begin{tcolorbox}[
        colback=white,
        colframe=black!70,
        boxrule=0.8pt,
        arc=2pt,
        title=\textbf{Scientific Types},
        fonttitle=\bfseries\small
    ]
    
    \begin{tabularx}{\textwidth}{@{}p{0.08\textwidth}XX@{}}
        \midrule
        \rowcolor{tbl_header}
        \textbf{Turn} & \textbf{User Query (Input)} & \textbf{Immediate Validation (State Assertion)} \\
        \midrule
        
        \multicolumn{3}{@{}l@{}}{\cellcolor{tbl_header!50}\textbf{\textit{Case: \texttt{dataframe\_merge} (Relational Logic)}}} \\
        \midrule
        \step{T1} & 
        ``Merge \texttt{df} and \texttt{df2} on product column. Store in \texttt{result\_df}.'' & 
        \validator{validate\_df\_merge} \newline
        $\bullet$ Assert \texttt{len(result\_df) == 3}. \newline
        $\bullet$ Assert column "supplier" exists. \\
        \midrule
        \step{T2} & 
        ``Update \texttt{result\_df} to keep only rows where supplier is 'SupA'.'' & 
        \validator{validate\_df\_merge\_filter} \newline
        $\bullet$ Assert \texttt{len(result\_df) == 2}. \newline
        $\bullet$ Logic: Keeps 'Phone' and 'Shirt'. \\
        \midrule
        \step{T3} & 
        ``Calculate the sum of prices in \texttt{result\_df}. Store in \texttt{result\_value}.'' & 
        \validator{validate\_df\_merge\_sum} \newline
        $\bullet$ Assert \texttt{result\_value == 550.0}. \newline
        $\bullet$ Logic: $500.0 + 50.0$. \\
        \midrule
        
        \multicolumn{3}{@{}l@{}}{\cellcolor{tbl_header!50}\textbf{\textit{Case: \texttt{dataframe\_pivot} (Structure Reshaping)}}} \\
        \midrule
        \step{T1} & 
        ``Create pivot table from \texttt{df\_sales}: region=rows, quarter=cols, sales=values.'' & 
        \validator{validate\_df\_pivot} \newline
        $\bullet$ Assert \texttt{result\_df.shape == (3, 2)}. \newline
        $\bullet$ Checks dimensions (3 regions, 2 quarters). \\
        \midrule
        \step{T2} & 
        ``Calculate total sum of all sales...'' & 
        \validator{validate\_df\_pivot\_sum} \newline
        $\bullet$ Assert \texttt{result\_value == 890}. \newline
        $\bullet$ Verifies data integrity post-pivot. \\
        \midrule
        \step{T3} & 
        ``Find which region has highest total sales (sum of Q1+Q2). Store sum...'' & 
        \validator{validate\_df\_pivot\_max\_region} \newline
        $\bullet$ Assert \texttt{result\_value == 380}. \newline
        $\bullet$ Logic: South ($200 + 180$). \\
        \midrule

        \multicolumn{3}{@{}l@{}}{\cellcolor{tbl_header!50}\textbf{\textit{Case: \texttt{ndarray\_reshape} (Tensor Manipulation)}}} \\
        \midrule
        \step{T1} & 
        ``Reshape array to shape (2, 4). Store in \texttt{result\_array}.'' & 
        \validator{validate\_array\_reshape} \newline
        $\bullet$ Assert \texttt{result\_array.shape == (2, 4)}. \newline
        $\bullet$ Checks memory layout transformation. \\
        \midrule
        \step{T2} & 
        ``Sum \texttt{result\_array} along \textbf{axis 1} (row sums).'' & 
        \validator{validate\_array\_sum\_axis} \newline
        $\bullet$ Assert result equals \texttt{[70, 48]}. \newline
        $\bullet$ Validates axis-wise reduction. \\
        \midrule
        \step{T3} & 
        ``Calculate the total sum of \texttt{result\_array}...'' & 
        \validator{validate\_array\_total} \newline
        $\bullet$ Assert \texttt{result\_value == 118}. \newline
        $\bullet$ Logic: $70 + 48$. \\
        \midrule
        
    \end{tabularx}
    \end{tcolorbox}
    \vspace{-0.5em}
    \caption{\label{fig:step_by_step_scientific} \textbf{Illustration of Scientific Types test cases.} This benchmark challenges the agent with high-dimensionality operations, including relational merges (\texttt{dataframe\_merge}), structural reshaping (\texttt{dataframe\_pivot}), and tensor axis manipulation (\texttt{ndarray\_reshape}), going beyond simple arithmetic.}
\end{figure*}
\subsection{Multi-variable Cases}
Since there are 5 tiers of variable numbers, we select the variable number = 20 to demonstrate our test case since different variable number shares similar patterns of test cases. Figure \ref{fig:multivar_cases_full} shows one example of test case where the agent is required to process 20 variables in 3 turns.
\begin{figure*}[!htbp]
    \centering
    \scriptsize 
    \renewcommand{\arraystretch}{1.3}
    
    \definecolor{tbl_header}{RGB}{240, 242, 245}
    \definecolor{turn_label}{RGB}{100, 100, 100}
    \definecolor{val_func}{RGB}{180, 0, 0}
    \definecolor{var_highlight}{RGB}{0, 50, 150}
    
    \providecommand{\step}[1]{\textcolor{turn_label}{\textbf{#1}}}
    \providecommand{\validator}[1]{\texttt{\textcolor{val_func}{#1}}}
    \newcommand{\varChange}[2]{\texttt{\textcolor{var_highlight}{#1}} $\to$ #2}

    \begin{tcolorbox}[
        colback=white,
        colframe=black!70,
        boxrule=0.8pt,
        arc=2pt,
        title=\textbf{Multi-Variable Management: Tracking 20 Concurrent States (Full Context)},
        fonttitle=\bfseries\small
    ]
    
    \begin{tabularx}{\textwidth}{@{}p{0.05\textwidth}Xp{0.28\textwidth}@{}}
        \midrule
        \rowcolor{tbl_header}
        \textbf{Turn} & \textbf{Complex User Query (Full Text)} & \textbf{State Verification (Partial View)} \\
        \midrule
        
        \multicolumn{3}{@{}l@{}}{\cellcolor{tbl_header!50}\textbf{\textit{Case: \texttt{startup\_journey} (20 Variables)}}} \\
        \midrule
        
        \step{T1} & 
        ``I'm documenting our startup TechStart. We're in the Software industry, led by CEO Alice Johnson, headquartered in San Francisco. We have 50 employees, founded in 2020, 1 office, 2 products. Revenue is \$5M (5000000) with 10\% profit margin (0.1), not public yet so no stock price or market cap. We're profitable and hiring but not international yet. Departments: ['Engineering', 'Sales', 'Marketing']. Locations: ['SF']. Financials: funding 10000000, round 'Series A'. Contacts: email 'info@techstart.com', phone '555-0100'.'' & 
        \validator{validate\_startup\_init} \newline
        $\bullet$ \varChange{employees}{50} \newline
        $\bullet$ \varChange{revenue}{5,000,000.0} \newline
        $\bullet$ \varChange{profit\_margin}{0.1} \newline
        $\bullet$ \varChange{public}{False} \newline
        $\bullet$ \varChange{stock\_price}{0.0} \textit{(Initial)} \\
        \midrule
        
        \step{T2} & 
        ``Big growth update! Set employees to 150, offices to 3, products to 5. Set revenue to \$15M (15000000), profit\_margin to 0.15. Set international to true. Append 'HR' and 'Finance' to departments. Append 'NYC' and 'London' to locations. Add 'valuation': 100000000 to financials while keeping existing entries. Add 'support': '555-0200' to contacts while keeping existing entries.'' & 
        \validator{validate\_startup\_growth} \newline
        $\bullet$ \varChange{employees}{150} \newline
        $\bullet$ \varChange{depts}{[..., 'HR', 'Finance']} \newline
        $\bullet$ \varChange{financials}{+\{'valuation': 100M\}} \newline
        $\bullet$ Assert \texttt{founded\_year == 2020} \textit{(Unchanged)} \\
        \midrule
        
        \step{T3} & 
        ``We're going public! Append ' Inc.' to company\_name. Set industry to 'Enterprise Software'. Set employees to 500, offices to 10, products to 10. Set revenue to \$50M (50000000), profit\_margin to 0.2, stock\_price to 25.0, market\_cap to \$500M (500000000). Set public to true. Append 'Legal' and 'IR' to departments. Append 'Tokyo' and 'Berlin' to locations. Add 'ipo': true to financials while keeping existing entries. Add 'ir': 'ir\@techstart.com' to contacts while keeping existing entries.'' & 
        \validator{validate\_startup\_ipo} \newline
        $\bullet$ \varChange{public}{True} \newline
        $\bullet$ \varChange{stock\_price}{25.0} \newline
        $\bullet$ \varChange{company\_name}{"TechStart Inc."} \newline
        $\bullet$ \varChange{market\_cap}{500,000,000.0} \\
        \midrule
    \end{tabularx}
    \end{tcolorbox}
    \vspace{-0.5em}
    \caption{\label{fig:multivar_cases_full} \textbf{High-Dimensional State Management (Full Transcript).} We present the raw input queries for the \texttt{startup\_journey} case. The high information density requires the agent to parse and update over 10 distinct variables (Integers, Floats, Strings, Lists, Dictionaries) in a single turn (e.g., T3) without hallucination or omitting details.}
\end{figure*}

\clearpage
\subsection{Multi-turn Cases}
These test cases evaluate the agent's capability to process sequential instructions and maintain state precision over long-horizon scenarios. Unlike single-turn tasks where information is self-contained, these scenarios require the agent to maintain persistent memory of the system's status, as subsequent queries depend on the outcome of previous actions. We categorize these multi-turn benchmarks into two domains: \textbf{Smart Home Control} and \textbf{Financial Account Management}.

\subsubsection{Smart Home}
In the \textbf{Smart Home} scenario, the agent acts as a central automation controller responsible for managing a suite of simulated IoT devices, including smart lighting, thermostats, motorized blinds, security cameras, and media players. 

This benchmark specifically targets two advanced capabilities in stateful management:

\begin{itemize}
    \item Users frequently issue relative commands rather than absolute ones (e.g., ``turn up the music \textit{more}'' or ``dim the lights \textit{a bit}''). To execute these correctly, the agent must recall the exact discrete level set in previous turns (e.g., incrementing volume from 'medium' to 'high') rather than resetting to a default value.
    
    \item The agent must dynamically adjust device states based on simulated environmental contexts (e.g., ``sunset'', ``motion detected'') and complex user-defined conditions (e.g., ``if the temperature drops below 10$^\circ$C, set heating to 22$^\circ$C'').
\end{itemize}

As illustrated in Figure \ref{fig:smarthome_multiturn}, the \texttt{weekend\_party} case spans a simulated 24-hour cycle. The agent must maintain a coherent environment state, transitioning from a quiet morning to a loud party and finally to a secure night mode, without drifting from the user's cumulative intent.
\begin{figure*}[!htbp]
    \centering
    \footnotesize
    \renewcommand{\arraystretch}{1.3}
    
    \definecolor{tbl_header}{RGB}{240, 242, 245}
    \definecolor{time_badge}{RGB}{0, 80, 150}   
    \definecolor{state_box}{RGB}{245, 245, 245}  
    \definecolor{val_func}{RGB}{180, 0, 0}       
    \definecolor{change_color}{RGB}{0, 100, 50}  

    \providecommand{\step}[1]{\textcolor{gray}{\textbf{#1}}}
    \providecommand{\validator}[1]{\texttt{\textcolor{val_func}{#1}}}
    \newcommand{\timestamp}[1]{\colorbox{time_badge}{\textcolor{white}{\textbf{#1}}}}
    \newcommand{\stateChange}[2]{#1 $\xrightarrow{\text{act}}$ \textcolor{change_color}{\textbf{#2}}}

    \begin{tcolorbox}[
        colback=white,
        colframe=black!70,
        boxrule=0.8pt,
        arc=2pt,
        title=\textbf{Multi-Turn Scenario: Smart Home "Weekend Party" (Selected Turns)},
        fonttitle=\bfseries\small
    ]
    
    \begin{tabularx}{\textwidth}{@{}p{0.12\textwidth}Xp{0.32\textwidth}@{}}
        \midrule
        \rowcolor{tbl_header}
        \textbf{Time / Turn} & \textbf{User Query (Intent \& Context)} & \textbf{State Evolution \& Validation} \\
        \midrule
        
        \step{Turn 3} \newline
        \timestamp{1:00 PM} & 
        ``Party prep! Guests arriving soon. Adjust thermostat for comfort, set music to medium, open blinds fully, make lights bright.'' & 
        \validator{validate\_party\_turn\_3} \newline
        $\bullet$ Music: OFF $\to$ \textbf{40\% (Medium)} \newline
        $\bullet$ Blinds: Closed $\to$ \textbf{100\% (Full)} \newline
        $\bullet$ Light: Dim $\to$ \textbf{80\% (Bright)} \\
        \midrule
        
        \step{Turn 5} \newline
        \timestamp{4:00 PM} & 
        ``Party mode! Full swing now. \textbf{Turn up the music} and make lights \textbf{very bright}. Verify camera is recording.'' & 
        \validator{validate\_party\_turn\_5} \newline
        $\bullet$ Music: 50\% $\to$ \textbf{60\% (Party)} \newline
        $\bullet$ Light: 80\% $\to$ \textbf{90\% (Very Bright)} \newline
        $\bullet$ Camera: Assert status == \textbf{Recording} \\
        \midrule
        
        \step{Turn 7} \newline
        \timestamp{7:00 PM} & 
        ``Evening party. \textbf{Close blinds completely}, set mood lighting... turn up music \textbf{more}.'' & 
        \validator{validate\_party\_turn\_7} \newline
        $\bullet$ Blinds: Partial $\to$ \textbf{0\% (Closed)} \newline
        $\bullet$ Music: 60\% $\to$ \textbf{70\% (Up More)} \newline
        $\bullet$ Light: 90\% $\to$ \textbf{60\% (Mood)} \\
        \midrule
        
        \step{Turn 10} \newline
        \timestamp{10:00 PM} & 
        ``Guests leaving. \textbf{Lower music more}, lock door, turn off bedroom light.'' & 
        \validator{validate\_party\_turn\_10} \newline
        $\bullet$ Music: 80\% $\to$ \textbf{< 60\% (Lowered)} \newline
        $\bullet$ Door: Unlocked $\to$ \textbf{Locked} \newline
        $\bullet$ Bed Light: ON $\to$ \textbf{OFF} \\
        \midrule

        \step{Turn 17} \newline
        \timestamp{Sun 10 AM} & 
        ``Lazy morning... Finally getting up. Turn on bedroom light, open blinds, raise thermostat.'' & 
        \validator{validate\_party\_turn\_17} \newline
        $\bullet$ \textit{Long-horizon consistency check} \newline
        $\bullet$ Thermostat: Eco (18) $\to$ \textbf{Comfort (21)} \newline
        $\bullet$ Blinds: Closed $\to$ \textbf{70\% (Open)} \\
        \midrule
        
    \end{tabularx}
    \end{tcolorbox}
    \vspace{-0.5em}
    \caption{\label{fig:smarthome_multiturn} \textbf{State persistence in long-horizon interactions.} We visualize 5 key moments from the 20-turn \texttt{weekend\_party} scenario. The agent must maintain a coherent environment state (lighting, temperature, security, audio) over a simulated 24-hour period. Crucially, it handles \textbf{relative instructions} (e.g., "turn up music", "lower music more") by tracking the exact discrete levels (e.g., Medium=40, Party=60) defined in the environment schema.}
\end{figure*}

\subsubsection{Financial Account}
The \textbf{Financial Account} benchmark evaluates the agent's capability to maintain \textbf{strict numerical integrity} and execute \textbf{state-dependent logic} within a banking ledger system. Unlike the relative adjustments in Smart Home, this domain demands exact integer arithmetic, where the agent must process a continuous stream of transactions, including deposits, interest applications, and loan amortizations, without cumulative drift.

This scenario imposes two constraints designed to stress-test the agent's reasoning stability:

\begin{itemize}
    \item Operations require strict integer truncation (e.g., calculating $20\%$ of $1105$ as $221$, not $221.0$). Since the output of each turn (e.g., current balance) serves as the immutable basis for subsequent calculations (e.g., compound interest), a single arithmetic error in early turns triggers a cascading failure, rendering the entire subsequent interaction trajectory incorrect.
    
    \item The agent must evaluate complex logic gates based on dynamic runtime states rather than static instructions. As demonstrated in the \texttt{carol\_debt\_paydown} case (Figure \ref{fig:financial_multiturn}), queries often involve comparative functions (e.g., ``pay the \textit{smaller} of 15\% of balance or 15\% of loan'') or threshold checks (e.g., upgrading to `premium` status only if net worth becomes positive). This requires the agent to retrieve, compare, and act upon multiple variable states simultaneously before executing a transaction.
\end{itemize}

\begin{figure*}[!htbp]
    \centering
    \footnotesize
    \renewcommand{\arraystretch}{1.3}
    
    \definecolor{tbl_header}{RGB}{235, 240, 245}
    \definecolor{logic_tag}{RGB}{120, 0, 120}    
    \definecolor{calc_val}{RGB}{0, 100, 50}      
    \definecolor{val_func}{RGB}{180, 0, 0}       
    \definecolor{turn_label}{RGB}{100, 100, 100}

    \providecommand{\step}[1]{\textcolor{turn_label}{\textbf{#1}}}
    \providecommand{\validator}[1]{\texttt{\textcolor{val_func}{#1}}}
    \newcommand{\logic}[1]{\textcolor{logic_tag}{\textbf{[Logic]}} \textit{#1}}
    \newcommand{\calc}[1]{\textcolor{calc_val}{\texttt{#1}}}

    \begin{tcolorbox}[
        colback=white,
        colframe=black!70,
        boxrule=0.8pt,
        arc=2pt,
        title=\textbf{Multi-Turn Scenario: Financial Account "Carol's Debt Paydown" (Numerical Precision)},
        fonttitle=\bfseries\small
    ]
    
    \begin{tabularx}{\textwidth}{@{}p{0.05\textwidth}Xp{0.35\textwidth}@{}}
        \midrule
        \rowcolor{tbl_header}
        \textbf{Turn} & \textbf{Conditional Query (Logic \& Math)} & \textbf{State Calculation \& Assertions} \\
        \midrule
        
        \step{T1} & 
        ``Initialize account... Name 'Carol', \textbf{Balance 500}, Status 'standard', Interest 8\% (Loan rate), \textbf{Loan 2000}.'' & 
        \validator{validate\_carol\_turn\_1} \newline
        $\bullet$ Balance: 500 \newline
        $\bullet$ Loan: 2000 \newline
        $\bullet$ Status: 'standard' \\
        \midrule
        
        \step{T2} & 
        ``Monthly loan interest due. Apply interest rate (8\%) to loan balance and add to debt.'' & 
        \validator{validate\_carol\_turn\_2} \newline
        $\bullet$ Interest = $2000 \times 0.08 = 160$ \newline
        $\bullet$ New Loan = $2000 + 160 = \calc{2160}$ \\
        \midrule
        
        \step{T4} & 
        ``Pay the \textbf{smaller of} 15\% of balance or 15\% of loan\_balance. Subtract from both.'' \newline
        \textit{(Context: T3 Paycheck +800 $\to$ Balance 1300)} & 
        \validator{validate\_carol\_turn\_4} \newline
        $\bullet$ \logic{IF} $\min(1300 \times .15, 2160 \times .15)$ \newline
        $\bullet$ Calc: $\min(195, 324) = 195$ \newline
        $\bullet$ New Loan = $2160 - 195 = \calc{1965}$ \\
        \midrule
        
        \step{T8} & 
        ``Pay the \textbf{larger of} 40\% of balance or 500 toward loan.'' \newline
        \textit{(Context: Balance grew to 1574 after T7)} & 
        \validator{validate\_carol\_turn\_8} \newline
        $\bullet$ \logic{Compare:} $1574 \times 0.4$ (629) vs 500 \newline
        $\bullet$ Action: Pay \calc{629} \newline
        $\bullet$ Verify exact integer subtraction. \\
        \midrule

        \step{T14} & 
        ``Check upgrade: \textbf{IF} loan\_balance < balance, upgrade status to 'premium'.'' \newline
        \textit{(Context: Loan reduced to 1172, Balance 1646)} & 
        \validator{validate\_carol\_turn\_14} \newline
        $\bullet$ \logic{Condition:} $1172 < 1646$ (True) \newline
        $\bullet$ Status $\to$ \textbf{'premium'} \newline
        $\bullet$ \textit{Triggers T15 bonus paycheck.} \\
        \midrule
        
        \step{T16} & 
        ``\textbf{IF} balance > loan\_balance, pay off entire loan. Otherwise pay 75\%...'' & 
        \validator{validate\_carol\_turn\_16} \newline
        $\bullet$ \logic{Action:} Payoff Condition Met. \newline
        $\bullet$ Loan $\to$ \calc{0.0} \newline
        $\bullet$ Balance reduced by remaining debt. \\
        \midrule
        
    \end{tabularx}
    \end{tcolorbox}
    \vspace{-0.5em}
    \caption{\label{fig:financial_multiturn} \textbf{Numerical precision and state-dependent reasoning.} In the \texttt{carol\_debt\_paydown} scenario, the agent must perform exact integer arithmetic while navigating complex logic gates (e.g., Turn 4's "smaller of", Turn 14's "net worth check"). A single miscalculation in early turns (e.g., T2 interest) would cascade, causing failures in subsequent logic checks (e.g., failing the T16 payoff condition), thus rigorously testing long-horizon numerical stability.}
\end{figure*}


\section{Stateful Runtime-Mediated Multi-Agent Coordination}
\label{app_multiagent}

The function-calling paradigm in CaveAgent introduces three key contributions for multi-agent coordination; Figure~\ref{Town} (shown at the start of the paper) illustrates an example. In this paper, we focus on qualitative analysis and provide case studies to facilitate understanding, leaving rigorous quantitative evaluation for future work. We introduce the high-level ideas below.

\subsection{Meta-Agent Runtime Control}
Sub-agents are injected as first-class objects into a meta-agent's runtime, enabling the meta-agent to programmatically access and manipulate child agent states through generated code. Rather than following predefined communication protocols, the meta-agent dynamically sets variables in sub-agent runtimes, triggers execution, and retrieves results, enabling adaptive pipeline construction, iterative refinement loops, and conditional branching based on intermediate states.

\subsection{State-Mediated Communication} Inter-agent data transfer bypasses message passing entirely. Agents communicate through direct runtime variable injection: the meta-agent retrieves objects from one agent's runtime and injects them into another's as native Python artifacts (DataFrames, trained models, statistical analyses), preserving type fidelity and method interfaces without serialization loss.

\subsection{Shared-Runtime Synchronization}
For peer-to-peer coordination, multiple agents can operate on a unified runtime instance, achieving implicit synchronization without explicit messaging. When one agent modifies a shared object, all peers perceive the change immediately through direct reference. New entities injected into the shared runtime become instantly discoverable, enabling collaborative manipulation of a unified world model with low coordination overhead.

\textbf{How the town simulation demonstrates this capability.} When the meta-agent modifies the weather state, all resident agents observe the change through direct attribute access; when a new location and manager are injected, existing agents can immediately query and interact with them.

Together, these patterns transform multi-agent systems from lossy text-based message exchange into typed, verifiable state flow, enabling automated validation of inter-agent handoffs and integration with downstream pipelines.

\subsection{AutoML Training Loop}
\label{app:automl}

\begin{figure}[!htbp]
    \centering
    \includegraphics[width=\linewidth]{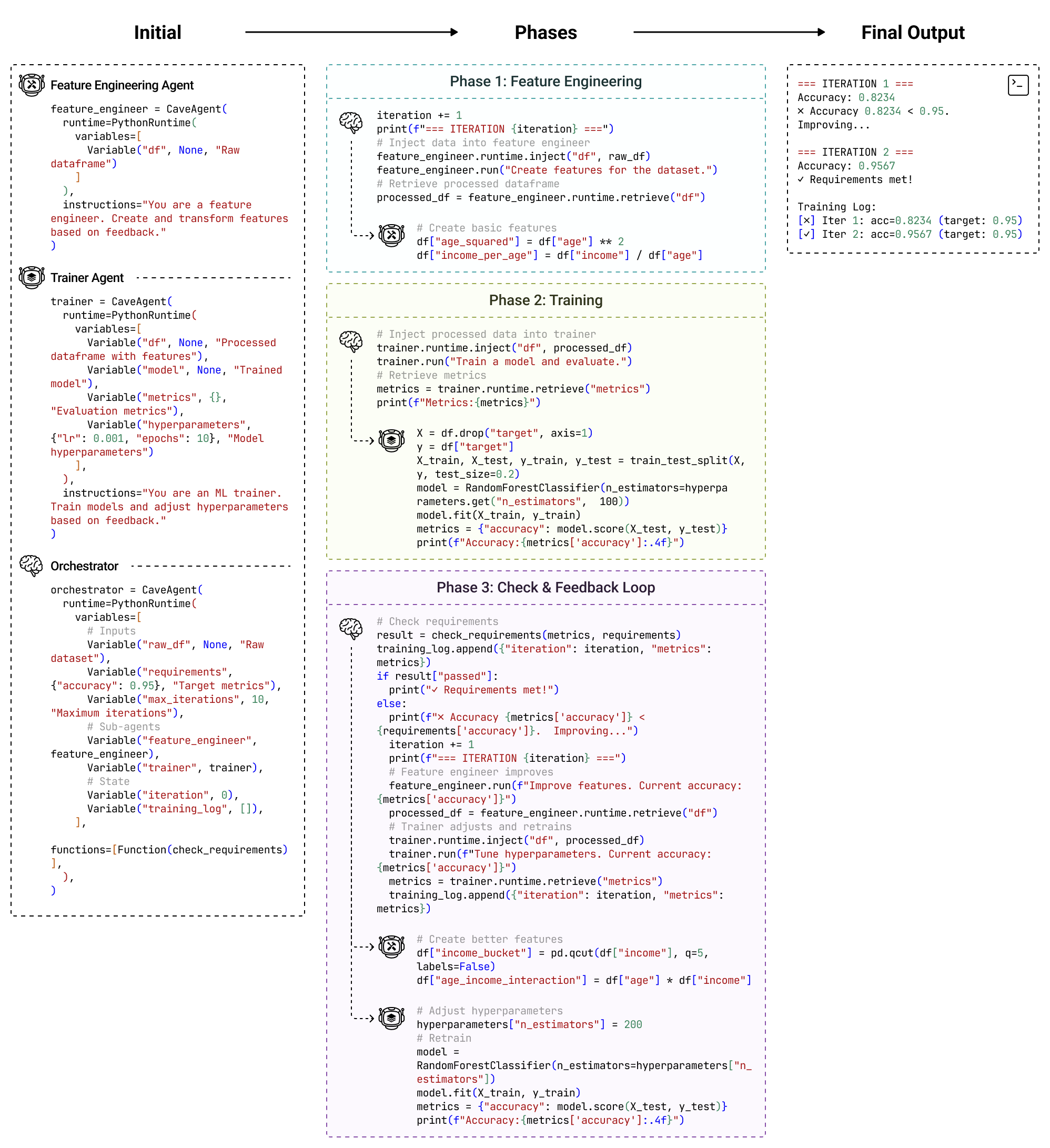}
    \caption{Multi-agent AutoML pipeline: an orchestrator coordinates feature engineering and training sub-agents through runtime variable injection. Native Python objects (DataFrames, trained models) flow losslessly between agents across iterative refinement cycles.}
    \label{fig:automl}
\end{figure}

We demonstrate CaveAgent's hierarchical agent coordination through an AutoML training loop where an orchestrator agent programmatically manages sub-agent runtimes (Figure~\ref{fig:automl}). The orchestrator injects raw data into a feature engineering agent's runtime via \texttt{inject()}, triggers execution, then retrieves the transformed DataFrame via \texttt{retrieve()} and injects it into a trainer agent's runtime, all as native Python objects without serialization. After training, the orchestrator extracts evaluation metrics directly from the trainer's runtime, validates against target requirements via a \texttt{check\_requirements()} function, and injects performance feedback back into both sub-agents for the next iteration. Crucially, sub-agents themselves are injected as variables into the orchestrator's runtime, enabling the orchestrator to dynamically access and manipulate their internal states through generated code. This iterative refinement loop continues until programmatic convergence criteria are met, demonstrating CaveAgent's unique capability for hierarchical multi-agent coordination with typed, bidirectional state flow and automated convergence verification.

\section{Detailed BFCL Benchmark Results}
\label{app:bfcl_detailed}
Table~\ref{tab:bfcl} shows the detailed per-run results of the BFCL benchmark.
\begin{table}[h]
\centering
\caption{Detailed Performance comparison on BFCL Benchmark (3 Runs). Data is presented in \textbf{score/total} format. The \colorbox{highlight}{Avg.} columns indicate the average overall percentage across 3 runs. \textbf{Bold} indicates CaveAgent outperforms Function Calling. Simp.\ = Simple Function, Mult.\ = Multiple Function, Para.\ = Parallel Function, P-M.\ = Parallel Multiple Function, Ov.\ = Overall.}
\label{tab:bfcl}
\small
\setlength{\tabcolsep}{1.2pt}
\renewcommand{\arraystretch}{1.2}
\begin{tabular}{@{}ll ccccc >{\columncolor{highlight}}c ccccc >{\columncolor{highlight}}c@{}}
\toprule
\multirow{2}{*}{\textbf{Model}} & \multirow{2}{*}{\textbf{Run}} & \multicolumn{6}{c}{\textbf{Function Calling}} & \multicolumn{6}{c}{\textbf{CaveAgent}} \\
\cmidrule(lr){3-8} \cmidrule(lr){9-14}
 &  & Simp. & Mult. & Para. & P-M. & Ov. & \textbf{Avg.(\%)} & Simp. & Mult. & Para. & P-M. & Ov. & \textbf{Avg.(\%)} \\
\midrule
\multicolumn{14}{l}{\textbf{\textit{Open Source}}} \\
\midrule
\multirow{3}{*}{\shortstack[l]{DeepSeek-V3.2\\{\tiny (685B)}}}
 & R1 & 354/400 & 183/200 & 175/200 & 159/200 & 871/1000 & & 382/400 & 192/200 & 185/200 & 178/200 & 937/1000 & \\
 & R2 & 353/400 & 185/200 & 167/200 & 159/200 & 864/1000 & & 386/400 & 193/200 & 184/200 & 178/200 & 941/1000 & \\
 & R3 & 360/400 & 185/200 & 173/200 & 154/200 & 872/1000 & \multirow{-3}{*}{86.9} & 384/400 & 192/200 & 186/200 & 180/200 & 942/1000 & \multirow{-3}{*}{\shortstack{\textbf{94.0} \\ \tiny \textcolor{teal}{(+7.1)}}} \\
\midrule
\multirow{3}{*}{\shortstack[l]{DeepSeek-V3.2\\{\tiny (w/o prompt)}}}
 & R1 & 312/400 & 162/200 & 33/200 & 26/200 & 533/1000 & & 382/400 & 192/200 & 185/200 & 178/200 & 937/1000 & \\
 & R2 & 316/400 & 162/200 & 29/200 & 23/200 & 530/1000 & & 386/400 & 193/200 & 184/200 & 178/200 & 941/1000 & \\
 & R3 & 314/400 & 161/200 & 35/200 & 21/200 & 531/1000 & \multirow{-3}{*}{53.1} & 384/400 & 192/200 & 186/200 & 180/200 & 942/1000 & \multirow{-3}{*}{\shortstack{\textbf{94.0} \\ \tiny \textcolor{teal}{(+40.9)}}} \\
\midrule
\multirow{3}{*}{\shortstack[l]{Qwen3-Coder\\{\tiny (30B)}}}
 & R1 & 381/400 & 185/200 & 166/200 & 167/200 & 899/1000 & & 386/400 & 191/200 & 187/200 & 180/200 & 944/1000 & \\
 & R2 & 381/400 & 185/200 & 166/200 & 167/200 & 899/1000 & & 387/400 & 189/200 & 189/200 & 181/200 & 946/1000 & \\
 & R3 & 381/400 & 185/200 & 164/200 & 167/200 & 897/1000 & \multirow{-3}{*}{89.8} & 386/400 & 190/200 & 189/200 & 178/200 & 943/1000 & \multirow{-3}{*}{\shortstack{\textbf{94.4} \\ \tiny \textcolor{teal}{(+4.6)}}} \\
\midrule
\multirow{3}{*}{\shortstack[l]{Kimi-K2-0905\\{\tiny (1000B)}}}
 & R1 & 372/400 & 183/200 & 170/200 & 168/200 & 893/1000 & & 387/400 & 191/200 & 186/200 & 187/200 & 951/1000 & \\
 & R2 & 368/400 & 181/200 & 167/200 & 171/200 & 887/1000 & & 381/400 & 189/200 & 188/200 & 186/200 & 944/1000 & \\
 & R3 & 373/400 & 185/200 & 173/200 & 165/200 & 896/1000 & \multirow{-3}{*}{89.2} & 379/400 & 191/200 & 188/200 & 187/200 & 945/1000 & \multirow{-3}{*}{\shortstack{\textbf{94.7} \\ \tiny \textcolor{teal}{(+5.5)}}} \\
\midrule
\multicolumn{14}{l}{\textbf{\textit{Closed Source}}} \\
\midrule
\multirow{3}{*}{\shortstack[l]{Claude\\Sonnet 4.5}}
 & R1 & 387/400 & 189/200 & 184/200 & 183/200 & 943/1000 & & 382/400 & 189/200 & 185/200 & 187/200 & 943/1000 & \\
 & R2 & 388/400 & 190/200 & 183/200 & 182/200 & 943/1000 & & 384/400 & 189/200 & 185/200 & 186/200 & 944/1000 & \\
 & R3 & 387/400 & 190/200 & 184/200 & 184/200 & 945/1000 & \multirow{-3}{*}{94.4} & 385/400 & 189/200 & 184/200 & 186/200 & 944/1000 & \multirow{-3}{*}{\shortstack{\textbf{94.4} \\ \tiny \textcolor{teal}{({\raise.17ex\hbox{$\scriptstyle\sim$}}0.0)}}} \\
\midrule
\multirow{3}{*}{GPT-5.1}
 & R1 & 366/400 & 183/200 & 174/200 & 173/200 & 896/1000 & & 367/400 & 186/200 & 172/200 & 176/200 & 901/1000 & \\
 & R2 & 367/400 & 186/200 & 173/200 & 169/200 & 895/1000 & & 354/400 & 184/200 & 174/200 & 174/200 & 886/1000 & \\
 & R3 & 367/400 & 185/200 & 174/200 & 172/200 & 898/1000 & \multirow{-3}{*}{89.6} & 356/400 & 180/200 & 170/200 & 175/200 & 881/1000 & \multirow{-3}{*}{\shortstack{88.9 \\ \tiny \textcolor{gray}{(-0.7)}}} \\
\midrule
\multirow{3}{*}{\shortstack[l]{Gemini 3\\Pro}}
 & R1 & 380/400 & 190/200 & 187/200 & 185/200 & 942/1000 & & 382/400 & 191/200 & 184/200 & 186/200 & 943/1000 & \\
 & R2 & 380/400 & 192/200 & 188/200 & 183/200 & 943/1000 & & 378/400 & 194/200 & 187/200 & 185/200 & 944/1000 & \\
 & R3 & 384/400 & 190/200 & 188/200 & 182/200 & 944/1000 & \multirow{-3}{*}{94.3} & 380/400 & 194/200 & 184/200 & 185/200 & 943/1000 & \multirow{-3}{*}{\shortstack{\textbf{94.3} \\ \tiny \textcolor{teal}{({\raise.17ex\hbox{$\scriptstyle\sim$}}0.0)}}} \\
\bottomrule
\end{tabular}
\end{table}

\section{Features}
\label{features_section}
\subsection{Case Analysis in Tau$^2$-bench}
\label{Additional_case:tau2}

To validate the architectural advantages of CaveAgent, we analyzed trajectory differences on the Tau$^2$-bench retail benchmark. CaveAgent achieved a 72.8\% success rate (83/114) compared to 62.3\% (71/114) for the baseline JSON agent (Kimi K2 backbone), yielding a 10.5\% improvement. We conducted a root cause analysis on the 24 tasks where CaveAgent succeeded but the baseline failed.

\subsubsection{Failure Taxonomy of the Baseline}
Baseline failures were categorized into five distinct patterns (Figure~\ref{fig:failure_dist}). The dominant failure mode (37.5\%) was \textit{Missing Critical Action}, where the agent retrieved necessary information but failed to execute the final operation (e.g., return, cancel). This was often coupled with \textit{Incomplete State Exploration} (16.7\%), where the agent heuristically queried subsets of data (e.g., checking only one recent order) rather than performing the exhaustive search required by the query.

\begin{figure}[!htbp]
    \centering
    \begin{tikzpicture}
        \begin{axis}[
            xbar,
            width=0.85\linewidth,
            height=6cm,
            symbolic y coords={Wrong Order Selection, Incomplete State Exploration, Missing Calculation, Wrong Arguments, Missing Critical Action},
            ytick=data,
            nodes near coords={\pgfmathprintnumber\pgfplotspointmeta\%},
            nodes near coords align={horizontal},
            xlabel={Percentage of Failure Cases (N=24)},
            xmin=0, xmax=45,
            y tick label style={text width=4cm, align=right},
            bar width=15pt,
            axis x line=bottom,
            axis y line=left,
            enlarge y limits=0.15,
            grid=major
        ]
        \addplot[fill=blue!40] coordinates {
            (16.7,Wrong Order Selection)
            (16.7,Incomplete State Exploration)
            (20.8,Missing Calculation)
            (25.0,Wrong Arguments)
            (37.5,Missing Critical Action)
        };
        \end{axis}
    \end{tikzpicture}
    \caption{Distribution of failure modes in baseline JSON agent trajectories. Note: Categories are non-exclusive as complex tasks may exhibit multiple failures.}
    \label{fig:failure_dist}
\end{figure}

\subsubsection{Architectural Advantages: Loops and Conditionals}
The analysis reveals that CaveAgent's improvements stem from its ability to generate programming constructs, specifically loops (used in 92\% of winning cases) and conditionals (83\%), which resolve the semantic gaps inherent in single-step function calling.

\paragraph{Exhaustive State Exploration via Loops.}
Tasks requiring global search (e.g., "return the order sent to Texas") baffled the baseline agent, which typically checked only 1--2 arbitrary orders. In contrast, CaveAgent generated \texttt{for}-loops to iterate through all user orders. For instance, in Task 26, the agent iterated through \texttt{user.orders}, checked \texttt{order.address.state} for "TX", and correctly identified the target order without hallucination.

\begin{lstlisting}[language=Python, caption={Snippet from Task 26 showing exhaustive search.}, label={lst:loop_example}]
# CaveAgent: Systematic iteration ensures no order is missed
for order_id in user_details.orders:
    order = get_order_details(order_id)
    if "TX" in order.address.state:
        return_delivered_order_items(order_id, ...)
\end{lstlisting}

\paragraph{Complex Conditional Logic.}
The baseline struggled with tasks involving fallback logic (e.g., "modify item, but if price > \$3000, cancel order"). In Task 90, the JSON agent ignored the price constraint and attempted modification regardless. CaveAgent successfully modeled this decision tree using explicit \texttt{if/else} blocks, checking variable states (\texttt{variant.price}) before execution.

\paragraph{Precise Attribute Reasoning.}
While JSON agents rely on the LLM's internal attention to compare values (often leading to errors like cancelling the wrong order in Task 59), CaveAgent offloads reasoning to the Python interpreter. By storing intermediate results (e.g., timestamps) in variables and using comparison functions (e.g., \texttt{min()}), CaveAgent ensured precise argument selection for actions requiring temporal or numerical comparisons.

\subsection{Smart Home}
\begin{figure}[!htbp]
    \centering
    \includegraphics[width=\linewidth]{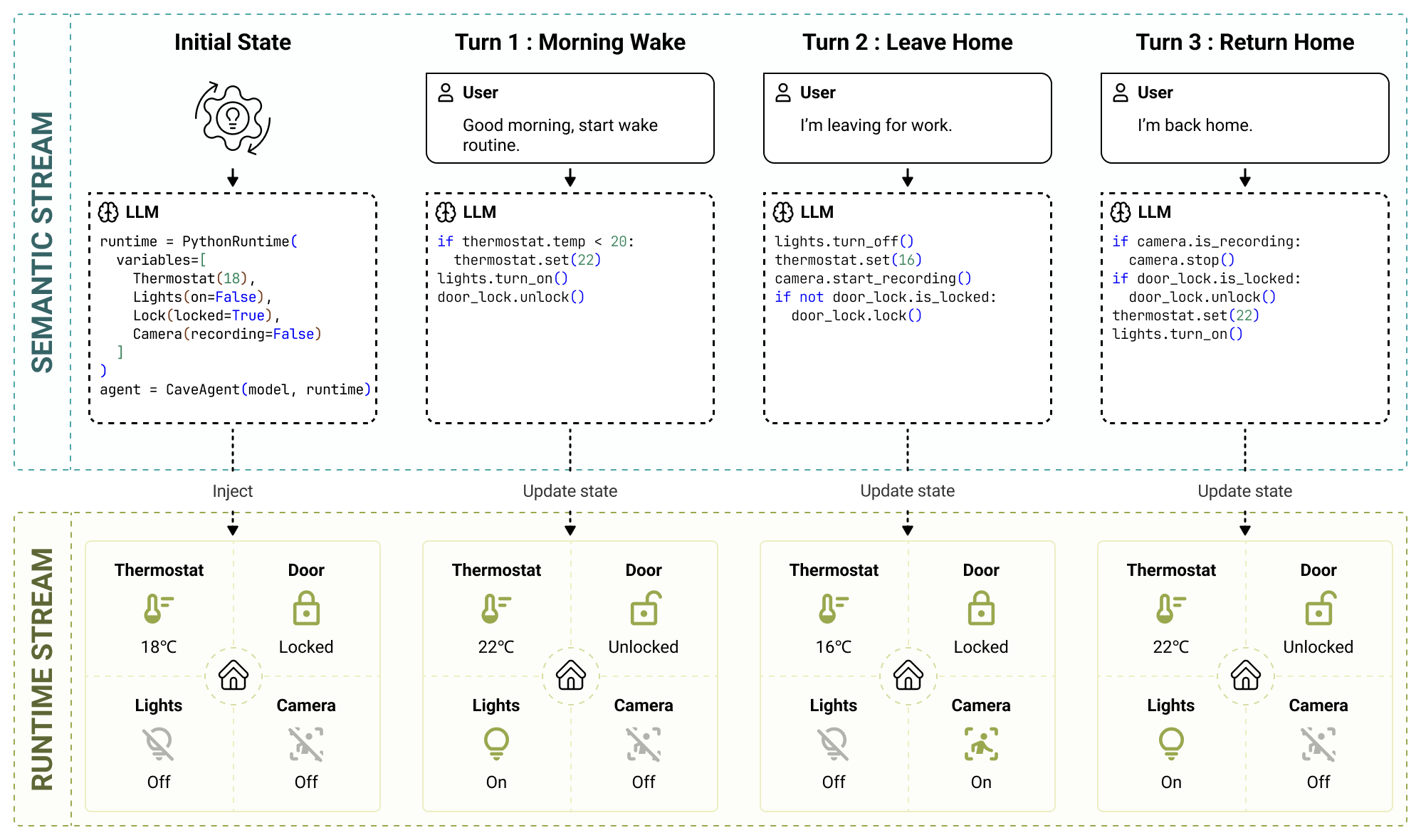}
    \caption{Demonstration of CaveAgent in a smart-home example: the Semantic Stream interacts with the Runtime Stream by generating code that manipulates stateful objects (variables) in the persistent runtime.}
    \label{fig:smart_home_demo}
\end{figure}

Figure~\ref{fig:smart_home_demo} illustrates the mechanistic advantage of CaveAgent through a toy smart-home example.
The architecture separates the \textit{Semantic Stream} (logic generation) from the \textit{Runtime Stream} (state storage).
This design enables two key capabilities absent in standard JSON agents:
\begin{itemize}
    \item \textbf{State Persistence:} Variables (e.g., \texttt{Thermostat}, \texttt{Door}) are initialized once and retain their state across multiple turns, eliminating the need to hallucinate or re-query context.
    \item \textbf{Control Flow Execution:} The agent generates executable Python code with conditionals \\(e.g., \texttt{if not door\_lock.is\_locked:}), allowing for precise, context-dependent state transitions rather than blind API execution.
\end{itemize}

\subsection{Geospatial Analysis}
\label{app:geospatial}

\begin{figure}[!htbp]
    \centering
    \includegraphics[width=\linewidth]{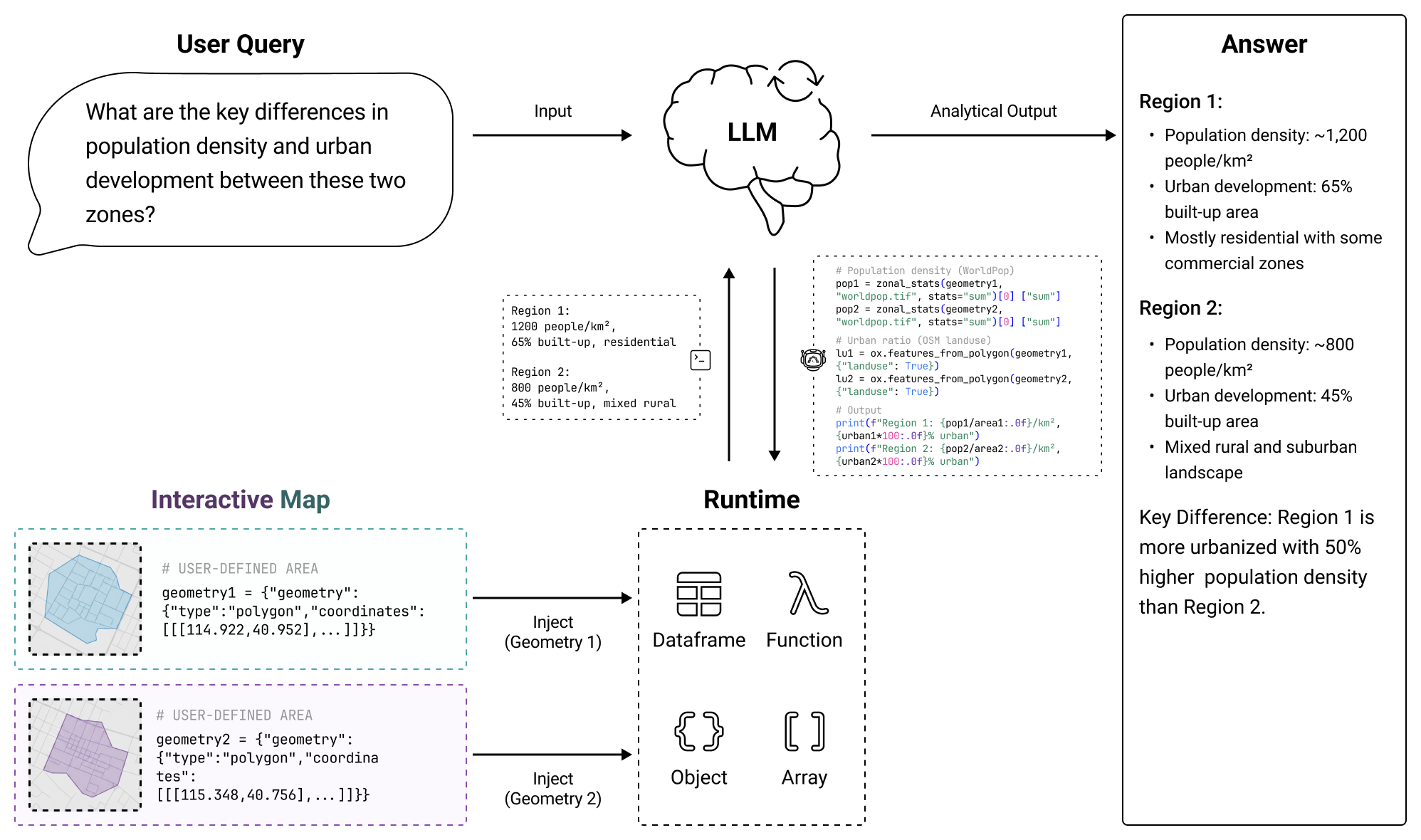}
    \caption{Geospatial Analysis: a user draws two regions on an interactive map. CaveAgent injects the GeoJSON polygons as Python variables and resolves the entire spatial query in a single turn with lossless data flow.}
    \label{fig:geospatial}
\end{figure}

To illustrate CaveAgent's advantages in domain-specific applications, we consider an urban planning scenario in which a user draws two arbitrary regions on an interactive map and queries the differences in population density and urban development between them. The interactive map converts user-drawn regions into GeoJSON polygon objects, variable-length coordinate arrays with high-precision floating-point pairs, which are injected directly into CaveAgent's persistent runtime as first-class Python variables. Upon receiving the natural language query, the LLM generates a single code block that references these injected geometries to perform zonal statistics against WorldPop raster data and extract land use features from OpenStreetMap via \texttt{osmnx}, chaining multiple domain-specific operations through native variable passing. Under the conventional JSON function calling paradigm, the same task would require at least five sequential LLM turns, querying population and land use statistics separately for each region before synthesizing text-serialized results, while also confronting the challenge of encoding complex polygon geometries as JSON string parameters, which risks truncation and introduces serialization overhead. CaveAgent resolves the entire query in a single turn with lossless data flow: geometries maintain full numerical precision, intermediate results (e.g., GeoDataFrames) persist as manipulable runtime objects, and the LLM synthesizes the final response from deterministic execution output (Figure~\ref{fig:geospatial}). This case study demonstrates CaveAgent's suitability for scientific and analytical domains where computation involves complex non-serializable data structures and precision-sensitive results.

\end{document}